\documentclass[10pt,twocolumn,twoside]{IEEEtran} 
\usepackage{amsmath,amsfonts}
\usepackage{algorithmic}
\usepackage{algorithm}
\usepackage{array}
\usepackage[caption=false,font=normalsize,labelfont=sf,textfont=sf]{subfig}
\usepackage{textcomp}
\usepackage{stfloats}
\usepackage{url}
\usepackage{verbatim}
\usepackage{graphicx}
\usepackage{cite}
\usepackage{hyperref}
\hypersetup{hidelinks,colorlinks=true,allcolors=black,pdfstartview=Fit,breaklinks=true}
\usepackage{amssymb}
\usepackage{enumitem}
\usepackage{color}
\definecolor{red}{RGB}{201, 47, 7}

\begin{document}

\title{Distributed Control within a Trapezoid Virtual Tube Containing Obstacles for Robotic Swarms Subject to Speed Constraints}
\author{Yan Gao, Chenggang Bai, Quan Quan
	\thanks{Yan Gao is with the Tianjin Key Laboratory of Intelligent Control of Electrical Equipment, Tiangong University, Tianjin 300387, China, and was with the School of Automation Science and Electrical Engineering, Beihang University, Beijing 100191, China (email: { buaa\_gaoyan@buaa.edu.cn}). Chenggang Bai and Quan Quan are with the School of Automation Science and Electrical Engineering, Beihang University, Beijing 100191, China (email: {bcg@buaa.edu.cn}; {qq\_buaa@buaa.edu.cn}).
	
	This work was supported by the National Natural Science Foundation of China under Grant 61973015 and 62303350, the Natural Science Foundation of Tianjin City under Grant 23JCQNJC00660.
	}}



\maketitle

\begin{abstract}
In our previous work, we design a trapezoid virtual tube to guide robotic swarms through narrow openings. This paper extends the application of the trapezoid virtual tube to the  situations where there are static obstacles inside and robots have strict speed constraints. We first propose a distributed swarm controller for the trapezoid virtual tube without obstacles and present the relationship between the trapezoid virtual tube and speed constraints. Then a switching logic for obstacle avoidance is proposed by dividing the trapezoid virtual tube containing static obstacles into several sub trapezoid virtual tubes without obstacles. Formal analyses and proofs are presented to demonstrate that all robots can pass through the trapezoid virtual tube safely. Besides, we validate the effectiveness of our method through numerical simulations and real experiments.
\end{abstract}


\section{Introduction}

\subsection{Background}

In recent years, there has been a growing interest in the development of robotic swarm systems. With their characteristics of convenience and efficiency, robotic swarm systems bring new potentialities in many complex tasks, such as search and rescue, goods transportation and military reconnaissance. A common scenario is that robots need to pass through some narrow openings in cluttered environments \cite{zhou2022swarm}, such as doorframes and windows, while avoiding collisions with obstacles and other robots \cite{soria2021predictive,hu2022decentralized,lavaei2022formal}. In this paper, we summarize this process as a ``passing-through problem''.

\subsection{Existing Strategies for Passing-through Problem}
To address the passing-through problem, methods in the literature can be classified as follows: formation, multi-robot trajectory planning, and control-based methods. Formation methods typically require each robot to maintain a predefined pose and perform transformation operations when necessary \cite{zhao2019bearing,zhang2021robust}. For the passing-through problem, the affine formation control method is a feasible choice as it has a better control effect on transformation operations \cite{xu2020affine}. However, the scalability and adaptability of the formation are limited, which makes it unsuitable for a large-scale robotic swarm system.

Multi-robot trajectory planning generates collision-free trajectories with higher-order continuity for all robots \cite{zhou2021ego}. For the passing-through problem, the concept of the safe flight corridor is widely used \cite{park2020online,park2022online,toumieh2022decentralized}. Based on digital maps that mark obstacles, a safe flight corridor is a series of obstacle-free areas with convex shapes and interconnections, thus forming a corridor in which unmanned aerial vehicles (UAVs) can fly freely. However, as the number of robots increases, the computational complexity and communication requirements may become unacceptable, making trajectory planning unfeasible \cite{kushleyev2013towards}.

\begin{figure}[!t]
	\centerline{\includegraphics[scale=0.7]{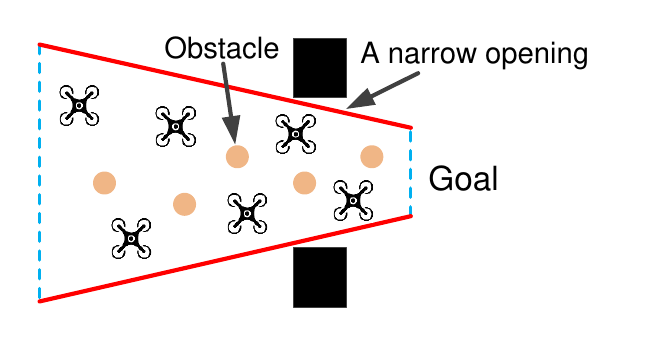}}
	\caption{A trapezoid virtual tube with some static obstacles inside is used to guide multiple multicopters through a narrow opening.}
	\label{First}
\end{figure}

Control-based methods use a simple controller to guide robots' movement based on local information \cite{panagou2015distributed,wang2017safety}. Among many control-based methods, the most popular is the artificial potential field (APF) method \cite{xie2022distributed}, which is particularly suitable for addressing multiple objectives simultaneously. The authors in \cite{chiang2015path,sharma2022priority,liu2018brain} use the APF method to guide robots through narrow openings or corridors. The main disadvantage of the APF method is its local minima problem \cite{guerra2016avoiding}. To solve this problem, various approaches have been designed, such as the harmonic potential function \cite{kim1992real,afzal2022model} and the Morse function \cite{rimon1990exact,panagou2015distributed}. Other control-based methods include vector field methods \cite{panagou2014motion}, control barrier function methods \cite{breeden2023robust}, flocking methods \cite{ibuki2020optimization}, etc. For the passing-through problem, control-based methods usually make the robotic swarm have a flexible structure instead of maintaining a fixed formation. Compared with trajectory planning, control-based methods have lower demands for computing and communication capabilities. Hence, control-based methods are most suitable for a large-scale robotic swarm system due to their simplicity and accessibility.

Inspired by vehicle traffic flow and fluid motion in nature, a \textit{trapezoid virtual tube} is proposed for the passing-through problem in our previous work \cite{gao2022distributed}. Similar to the safe flight corridor, the trapezoid virtual tube creates a safety zone in which robots do not need to consider obstacles, but rather only need to avoid collisions with each other and the virtual tube boundary. For the trapezoid virtual tube, we summarize two main problems: \textit{trapezoid virtual tube planning problem} and \textit{trapezoid virtual tube passing-through control problem}. The former can be solved using traditional path planning algorithms \cite{wang2020neural} or ``teach-and-repeat" systems \cite{mattamala2022efficient}. Our previous work \cite{gao2022distributed} focuses on the latter and proposes a distributed swarm controller.

\subsection{Motivation and Contribution}

In our previous work \cite{gao2022distributed}, the proposed swarm controller can be seen as a modified type of the APF method, whose output is the velocity commands of robots. The local minima problem for a robotic swarm is that the speeds of all robots are kept at zero when the robots have not reached their goal points. In \cite{gao2022distributed}, we solve this problem by introducing the single panel method \cite{kim1992real} to the swarm controller. However, in many cases, only the local minima avoidance is not enough. A common problem is that robots have minimum speed constraints \cite{wang2019motion}, which require the speeds of all robots be greater than a constant larger than zero. For the APF-based methods, an important feature is that the direction of the velocity command plays an important role in the safety and convergence analysis, while the norm of the velocity command can be scaled freely due to its lower importance \cite{gao2023non}. Thus, the local minima avoidance requires that the speeds of all robots cannot be zero at the same time, and the minimum speed constraint requires that the speed of any robot cannot be zero. If the velocity command is a zero vector, it cannot be scaled to meet the minimum speed constraint as the zero vector has no direction. We can conclude that the minimum speed constraint is more challenging for the swarm controller than the local minima problem. The minimum speed constraint is particularly necessary for the controller design of fixed-wing UAVs. To generate enough lifting forces, fixed-wing UAVs must keep their speeds greater than a \emph{stall speed} \cite{schmitt2014collision}. The consideration of the minimum speed constraint is also meaningful for other types of robots. For example, when multiple ground mobile robots are moving inside a trapezoid virtual tube, it is desired that the traffic efficiency is always larger than a specific value, which is obviously related to the minimum speed. In our previous work \cite{gao2022distributed}, only the maximum speed constraint is considered in the swarm controller. Hence, the minimum speed problem limits the application range of the trapezoid virtual tube.

Besides the problem of the minimum speed constraint, another problem in our previous work \cite{gao2022distributed} is that no obstacle is allowed to be inside the trapezoid virtual tube. It is obvious that this requirement is too ideal in real practice. There may exist many tiny static obstacles in cluttered environments \cite{xue2020tiny}. If these obstacles are all directly excluded from the virtual tube, the virtual tube planning may become unfeasible. In addition, during the passing-through process, some robots may have faults and be unable to continue moving. In this case, these robots may stay still in place. For other robots, these robots with faults can be considered static obstacles. A simple idea for obstacle avoidance is just introducing an obstacle avoidance function into the final potential field function \cite{kim1992real}. However, such a method may bring inevitable local minima problems. Besides, when the minimum speed constraint is also considered, the local minima problem in this paper is actually transferred into a dynamic obstacle avoidance problem, which is not easy to solve by the traditional APF method and will be described in detail in the following sections. 

Faced with the two issues introduced above, the problem of guiding robotic swarms through a trapezoid virtual tube containing obstacles subject to speed constraints is proposed and then solved in this paper. A distributed swarm controller is designed based on the single integrator model. Unlike the controller in \cite{gao2022distributed}, the robot's speed is always between the maximum speed and the minimum speed. We also present the relationship between the trapezoid virtual tube and the minimum speed constraint, which is unnecessary in our previous work \cite{gao2022distributed}. For obstacle avoidance inside the virtual tube, we partition the original trapezoid virtual tube that contains obstacles into multiple sub trapezoid virtual tubes without obstacles inside. Then, we design a switching logic for robots to pass through these sub trapezoid virtual tubes.
It should be noted that several single trapezoid virtual tubes can form a \emph{connected quadrangle virtual tube} as stated in \cite{gao2022distributed}. 
The switching logic proposed for the connected quadrangle virtual tube in \cite{gao2022distributed} can be utilized in this paper as well. Due to space limitations, this paper only concentrates on the trapezoid virtual tube. The main contributions of this paper are summarized as follows: 

\begin{itemize}[leftmargin=0.3cm]
	\item Based on our previous work \cite{gao2022distributed}, a distributed swarm controller is designed to guide multiple robots through the trapezoid virtual tube. The proposed controller always satisfies the maximum and minimum speed constraints, which extends the application range of the trapezoid virtual tube.

	\item The relationship between the trapezoid virtual tube and the minimum speed constraint is first proposed. We provide a sufficient condition for the trapezoid virtual tube to ensure the safety. No restriction is necessary if we only consider the maximum speed constraint. The proposed relationship also guides the planning of the trapezoid virtual tube.

	\item 
	Different from \cite{gao2022distributed}, the static obstacle avoidance inside the trapezoid virtual tube is achieved in this paper. To avoid the difficulties of convergence analysis caused by directly introducing repulsive potential functions for obstacles, we partition a trapezoid virtual tube that contains obstacles into multiple sub trapezoid virtual tubes without obstacles. Such a method can also adjust the obstacle avoidance behavior by parameter tuning.
	
\end{itemize} 


\section{Problem Formulation}

\subsection{Robot Model}
\subsubsection{Single Integrator Model}
In this paper, the robotic swarm consists of $M$ robots in $\mathbb{R}^2$. It is assumed that the $i$th robot is modeled as a single integrator
\begin{equation}
	\dot{\mathbf{{p}}}_{i} =\mathbf{v}_{\text{c},i},  \label{Single}
\end{equation}
where $\mathbf{v}_{\text{c},i}\in {{\mathbb{R}}^{2}}$, $\mathbf{p}_{i}\in {{\mathbb{R}}^{2}}$ are the velocity command and position. It is considered that all robots have speed constraints. As the robot's velocity and its velocity command are always the same in the single integrator model, here $\mathbf{v}_{\text{c},i}$ has such speed constraints as
\begin{equation}
	v_{\text{min}} \leq \left \Vert \mathbf{v}_{\text{c},i} \right \Vert \leq v_ {\text{max}}, \label{vminvmax}
\end{equation}
where $i=1,2,\cdots,M$, $v_{\text{min}}$ and $v_{\text{max}}$ are the minimum and the maximum permitted speeds, respectively. When robots are fixed-wing UAVs, $v_{\text{min}}$ is also called \emph{stall speed}.

\textbf{Remark 1}. In the case where a robot is modeled as a single integrator, such as multicopters and helicopters, we can use $\mathbf{v}_{\text{c},i}$ directly to control the robot. However, for more complex robot models, such as the double integrator or nonholonomic models, additional control laws are required.

\subsubsection{Two Areas around a Robot}
At the time $t>0$, the \emph{safety area} of the $i$th robot is defined as 
$
	\mathcal{S}_{i}\left(t\right)=\left \{ \mathbf{x}\in {{\mathbb{R}}^{2}}: \left \Vert \mathbf{x}-\mathbf{p}_{i}\left(t\right)\right \Vert \leq r_{\text{s}} \right \} ,
$
where $r_{\text{s}}>0$ is the \emph{safety radius} \cite{gao2022distributed}. For all robots to avoid conflict with each other, it is required that $\mathcal{S}_{i}\cap \mathcal{S}_{j}=\varnothing$,
where $i\neq j,i,j=1,\cdots , M$. The $i$th robot's \emph{avoidance area} is defined as
$
	\mathcal{A}_{i}\left(t\right)=\left \{ \mathbf{x}\in {{\mathbb{R}}^{2}}:\left
	\Vert \mathbf{x}- \mathbf{p}_{i}\left(t\right)\right \Vert \leq r_{\text{a}}
	\right \},
$
where $r_{\text{a}}>0$ is the \emph{avoidance radius}. 
It is required that
$r_{\text{a}}>r_{\text{s}}$. We define $\mathcal{N}_{\text{m},i}$ as the set of all IDs of other robots whose safety areas intersect the avoidance area of the $i$th robot, namely
$
	\mathcal{N}_{\text{m},i}=\left \{ j: \mathcal{A}_{i} \cap \mathcal{S}_{j}
	\neq \varnothing,j\neq i\right \}.
$

\subsection{Trapezoid Virtual Tube Model}


\begin{figure}[!t]
	\centering
	\includegraphics[scale=1.1]{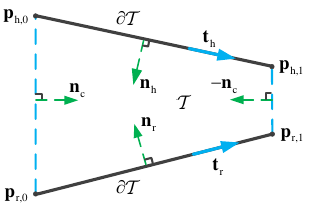}
	\caption{Trapezoid virtual tube.}
	\label{AllTube}
\end{figure}

As shown in Figure \ref{AllTube}(a), a trapezoid virtual tube $\mathcal{T}\left({{\mathbf{p}}_{\text{h},0},{\mathbf{p}}_{\text{h},1},{\mathbf{p}}_{\text{r},0},{\mathbf{p}}_{\text{r},1}}\right) $ is located in a horizontal plane ${\mathbb{R}}^{2}$. Points ${{\mathbf{p}}_{\text{h},0},{\mathbf{p}}_{\text{h},1},{\mathbf{p}}_{\text{r},0},{\mathbf{p}}_{\text{r},1}}\in {{\mathbb{R}}^{2}}$ are four vertices. The line segments $\left[ \mathbf{p}_{\text{h},0},{\mathbf{p}}_{\text{r},0}\right] $ and $\left[ \mathbf{p}_{\text{h},1},{\mathbf{p}}_{\text{r},1}\right] $ are two parallel bases. The line segment $\left[ \mathbf{p}_{\text{h},1},{\mathbf{p}}_{\text{r},1}\right] $ is also called the \emph{finishing line}. The line segments $\left[ \mathbf{p}_{\text{h},0},{\mathbf{p}}_{\text{h},1} \right]$ and $\left[ \mathbf{p}_{\text{r},0},{\mathbf{p}}_{\text{r},1}\right] $ are two legs, which are called \emph{left boundary} and \emph{right boundary}, respectively. 
Mathematically, the trapezoid virtual tube is represented as 
\begin{align*}
	\mathcal{T}=&\left \{ \mathbf{x}\in {{\mathbb{R}}^{2}} :
	\mathbf{n}_{\text{h}}^{\text{T}}\left( \mathbf{x}-{{\mathbf{p}}_{\text{h},1}}
	\right) \geq 0,\mathbf{n}_{\text{r}}^{\text{T}}\left( \mathbf{x}-{{\mathbf{p}}_{\text{r},1}}\right) \geq 0,\right.\\ 
	&\left.-\mathbf{n}_{\text{c}}^{\text{T}}\left( \mathbf{
		x}-{{\mathbf{p}}_{\text{r},1}}\right) \geq 0,\mathbf{n}_{\text{c}}^{\text{T}
	}\left( \mathbf{x}-{{\mathbf{p}}_{\text{r},0}}\right) \geq 0\right \}.
\end{align*}
The tube boundary of $\mathcal{T}$ is represented as
\begin{align*}
	\partial \mathcal{T}=\left \{ \mathbf{x}\in \mathcal{T}: 
	\mathbf{n}_{\text{h}}^{\text{T}}\left( \mathbf{x}-{{\mathbf{p}}_{\text{h},1}}
	\right) =0\cup \mathbf{n}_{\text{r}}^{\text{T}}\left( \mathbf{x}-{{\mathbf{p}}_{\text{r},1}}\right) =0\right \},
\end{align*}
where $\mathbf{n}_{\text{h}}$ and $\mathbf{n}_{\text{r}}$ are unit vectors in $\mathbb{R}^2$, and they are independent of the unit vector $\mathbf{n}_{\text{c}} \in \mathbb{R}^2$.
Define two unit vectors as $\mathbf{t}_{\text{h}}=\frac{\mathbf{p}_{\text{h},1}-\mathbf{p}_{\text{h},0}}{\left \Vert \mathbf{p}_{\text{h},1}-\mathbf{p}_{\text{h},0} \right \Vert}$, $\mathbf{t}_{\text{r}}=\frac{\mathbf{p}_{\text{r},1}-\mathbf{p}_{\text{r},0}}{\left \Vert \mathbf{p}_{\text{r},1}-\mathbf{p}_{\text{r},0} \right \Vert}$. 
And we have
$\mathbf{n}_{\text{h}}^{\text{T}}\mathbf{t}_{\text{h}} =0$, $\mathbf{n}_{\text{r}}^{\text{T}}\mathbf{t}_{\text{r}}=0 $, $\mathbf{t}_{\text{h}}^{\text{T}}\mathbf{n}_{\text{c}} >0$, $\mathbf{t}_{\text{r}}^{\text{T}}\mathbf{n}_{\text{c}} >0$.

\begin{figure}[!t]
	\centering
	\includegraphics[width=0.8\columnwidth]{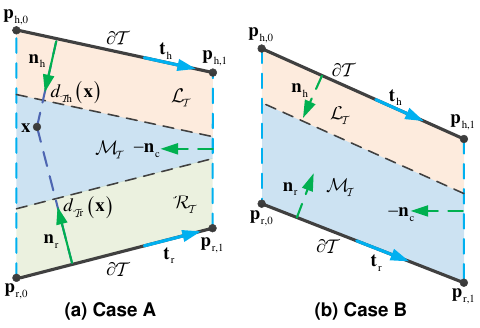}
	\caption{Three areas inside the trapezoid virtual tube. In plot (b), there exists $\mathcal{R}_\mathcal{T}=\varnothing$.}
	\label{MLR}
\end{figure}

When $\mathbf{x} \in \mathcal{T}$, two errors are defined as the distance between $\mathbf{x}$ and the tube boundary, namely
\begin{align}
	d_{\mathcal{T}\text{h}}\left(\mathbf{x}\right)&= \mathbf{n}_\text{h}^{\text{T}}\left(\mathbf{x}-{\mathbf{p}}_{\text{h},1}\right) \text{ if } \mathbf{x} \in \mathcal{T},  \label{dtl} \\
	d_{\mathcal{T}\text{r}}\left(\mathbf{x}\right)&= \mathbf{n}_\text{r}^{\text{T}}\left(\mathbf{x}-{\mathbf{p}}_{\text{r},1}\right) \text{ if } \mathbf{x} \in  \mathcal{T}.  \label{dtr}
\end{align}
As shown in Figure \ref{MLR}, the trapezoid virtual tube $\mathcal{T}$ is divided into three areas, namely the \emph{middle area} $\mathcal{M}_\mathcal{T}$,  the \emph{left area} $\mathcal{L}_\mathcal{T}$ and the \emph{right area} $\mathcal{R}_\mathcal{T}$, which are defined as
\begin{align*}
	\mathcal{M}_\mathcal{T}&=
	\begin{cases}
		\begin{aligned}
			\left \{ \mathbf{x}\in \mathcal{T}: d_{\mathcal{T}\text{h}}\left(\mathbf{x}\right)\geq k_\text{t} r_\text{a},\right.\\	\left. d_{\mathcal{T}\text{r}}\left(\mathbf{x}\right)\geq k_\text{t} r_\text{a}\right\}
		\end{aligned}
		& 	\text{if } \mathbf{n}_{\text{c}}^{\text{T}}\mathbf{n}_\text{h}<0,\mathbf{n}_{\text{c}}^{\text{T}}\mathbf{n}_\text{r}<0 \\
		\left \{ \mathbf{x}\in \mathcal{T}: d_{\mathcal{T}\text{h}}\left(\mathbf{x}\right)\geq k_\text{t} r_\text{a}\right\}
		& 	\text{if } \mathbf{n}_{\text{c}}^{\text{T}}\mathbf{n}_\text{h}<0,\mathbf{n}_{\text{c}}^{\text{T}}\mathbf{n}_\text{r}\geq0 \\
		\left \{ \mathbf{x}\in \mathcal{T}: d_{\mathcal{T}\text{r}}\left(\mathbf{x}\right)\geq k_\text{t} r_\text{a}\right\}
		& 	\text{if } \mathbf{n}_{\text{c}}^{\text{T}}\mathbf{n}_\text{h}\geq0,\mathbf{n}_{\text{c}}^{\text{T}}\mathbf{n}_\text{r}<0 \\
		\mathcal{T}
		& 	\text{if } \mathbf{n}_{\text{c}}^{\text{T}}\mathbf{n}_\text{h}\geq0,\mathbf{n}_{\text{c}}^{\text{T}}\mathbf{n}_\text{r}\geq0 \\
	\end{cases},\\
	\mathcal{L}_\mathcal{T}&=
	\begin{cases}
		\left \{ \mathbf{x}\in \mathcal{T}: d_{\mathcal{T}\text{h}}\left(\mathbf{x}\right)< k_\text{t} r_\text{a}\right\}
		& 	\text{if } \mathbf{n}_{\text{c}}^{\text{T}}\mathbf{n}_\text{h}<0\\
		\varnothing & 	\text{if } \mathbf{n}_{\text{c}}^{\text{T}}\mathbf{n}_\text{h}\geq 0
	\end{cases},\\
	\mathcal{R}_\mathcal{T}&=
	\begin{cases}
		\left \{ \mathbf{x}\in \mathcal{T}: d_{\mathcal{T}\text{r}}\left(\mathbf{x}\right)< k_\text{t} r_\text{a}\right\}
		& 	\text{if } \mathbf{n}_{\text{c}}^{\text{T}}\mathbf{n}_\text{r}<0\\
		\varnothing & 	\text{if } \mathbf{n}_{\text{c}}^{\text{T}}\mathbf{n}_\text{r}\geq 0
	\end{cases},
\end{align*}
where $k_\text{t}\geq 1$.  If $\mathcal{L}_\mathcal{T} \cap \mathcal{R}_\mathcal{T}=\varnothing$ and $\mathcal{M}_\mathcal{T}\neq\varnothing$, it can be obtained that 
$
	 \mathcal{T}=\mathcal{M}_\mathcal{T}\cap \mathcal{L}_\mathcal{T}\cap \mathcal{R}_\mathcal{T}. 
$

\textbf{Remark 2}. The trapezoid virtual tube is previously known in this paper. Due to space limitations, how to plan and generate the trapezoid virtual tube in a cluttered environment is not the focus of this paper. 

\subsection{Problem Formulation}

There are $P$ static obstacles inside the trapezoid virtual tube $\mathcal{T}$. The \emph{occupied area} of the $k$th obstacle is given by
\begin{equation}
	\mathcal{O}_{k}=\left \{ \mathbf{x}\in {{\mathbb{R}}^{2}}: \left \Vert \mathbf{x}-\mathbf{p}_{\text{o},k}\right \Vert \leq r_{\text{o},k} \right \} , \label{obstacle}
\end{equation}
where $r_{\text{o},k}>0$ is the radius, and $\mathbf{p}_{\text{o},k} \in \mathbb{R}^2$ is the center position, $k=1,\cdots ,P$. Then, some additional assumptions are proposed.

\textbf{Assumption 1}. The initial conditions of the robots satisfy $\mathcal{S}_{i}\left( 0\right) \subset \mathcal{T}$, $\mathcal{S}_{i}\left( 0\right)\cap \mathcal{S}_{j}\left( 0\right)=\varnothing$, $\mathcal{S}_{i}\left( 0\right)\cap \mathcal{O}_k=\varnothing$, where $i\neq j,i,j=1,\cdots ,M, k=1,\cdots ,P$.

\textbf{Assumption 2}.  Upon reaching the finishing line of $\mathcal{T}$, the robot will exit the trapezoid virtual tube and will not affect the movement of other robots behind it. Given ${\epsilon }_{\text{0}}>0,$ the $i$th robot arrives near the finishing line $\left[ {{\mathbf{p}}_{\text{r,}1},{\mathbf{p}}_{\text{h,}1}}\right]$ of $\mathcal{T}$ if 
\begin{align}
	-\mathbf{n}_{\text{c}}^{\text{T}}\left( \mathbf{p}_{i}-{{\mathbf{p}}_{\text{r,}1}}\right) \leq {\epsilon }_{\text{0}}. \label{TraCondition}
\end{align}

\textbf{Assumption 3}. The trapezoid virtual tube is long enough and wide enough to contain at least one robot. Mathematically, it is required that $\mathcal{L}_\mathcal{T} \cap \mathcal{R}_\mathcal{T}=\varnothing$, $\mathcal{M}_\mathcal{T}\neq\varnothing$, $ \mathbf{n}_{\text{c}}^{\text{T}}\left({\mathbf{p}}_{\text{r,}1}-{\mathbf{p}}_{\text{r,}0}\right)>2r_\text{a}$.

\textbf{Assumption 4}. The $i$th robot has the ability to acquire relative position information of other robots belonging to $\mathcal{N}_{\text{m},i}$ through direct measurement or network sharing. This information is represented as $\mathbf{{p}}_i-\mathbf{{p}}_j \in \mathbb{R}^2, j \in \mathcal{N}_{\text{m},i}$.

The main problem is proposed as follows.

\begin{itemize}[leftmargin=0.3cm]
	\item \textbf{Distributed passing-through control problem within a trapezoid virtual tube subject to speed constraints}. Under \textit{Assumptions 1-4} and the speed constraints \eqref{vminvmax}, design the velocity command $\mathbf{v}_{\text{c},i}$ to guide all robots to pass through $\left[ {{\mathbf{p}}_{\text{r,}1},{\mathbf{p}}_{\text{h,}1}}\right]$ of $\mathcal{T}$, meanwhile avoiding conflict with each other  ($\mathcal{S}_{i}\left(t\right)\cap  \mathcal{S}_{j}\left(t\right)=\varnothing $), avoiding collision with obstacles  ($\mathcal{S}_{i}\left(t\right)\cap  \mathcal{O}_{k}=\varnothing $) and staying within the trapezoid virtual tube ($\mathcal{S}_{i}\left(t\right)\cap \partial\mathcal{T}=\varnothing$), 
	where  $i,j=1,\cdots ,M,i \neq j,k=1,\cdots ,P,t>0$.
\end{itemize}

\section{Vector Field Design within Trapezoid Virtual Tube with No Obstacle Considered}
\subsection{Preliminaries}
\subsubsection{Vector Saturation Function}
For the swarm controller design, a vector saturation function is designed as \cite{gao2022distributed}
\begin{equation*}
	\mathbf{u}=\text{sat}\left(\mathbf{x},{a}\right) ={{\kappa}_{{a}}}\left(\mathbf{x}\right)\mathbf{x},
\end{equation*}
where $\mathbf{x}\in {{\mathbb{R}}^{2}}$ is the original vector, $\mathbf{u}\in {{\mathbb{R}}^{2}}$ is the saturated vector, $a>0$ is the maximum norm of $\mathbf{u}$, and
\begin{align*}
	\text{sat}\left( \mathbf{x},{a}\right) =
	\begin{cases}
		\mathbf{x}  & 	\left \Vert \mathbf{x}\right \Vert \leq {a}\\ 
		{a}\frac{\mathbf{x}}{\left \Vert \mathbf{x}\right \Vert } & \left \Vert \mathbf{x}\right \Vert  >{a}
	\end{cases},\ \ 
	{{\kappa }_{{a}}}\left( \mathbf{x}\right) =
	\begin{cases}
		1 & \left \Vert \mathbf{x}\right \Vert \leq {a}\\ 
		\frac{{a}}{\left \Vert \mathbf{x}\right \Vert } & 	\left \Vert \mathbf{x}\right \Vert >{a}
	\end{cases}.
\end{align*}
We have $0<{{\kappa }_{{a}}}\left( \mathbf{x}\right)
\leq 1$. For brevity, we will refer to ${{\kappa }_{{a}}}\left( \mathbf{x}\right)$ as ${{\kappa }_{{a}}}$. Besides, $\mathbf{x}$ and  $\mathbf{u}$ always have the same direction. In this work, the vector saturation function is used for the swarm controller to satisfy the speed constraints of the robots.

\subsubsection{Two Smooth Functions and a Nominal Lyapunov-Like Barrier Function}
The first smooth function is designed as \cite{gao2022distributed,panagou2015distributed}
\begin{equation*}
	\sigma \left(x,d_{1},d_{2}\right)=\left \{ 
	\begin{array}{c}
		1 \\ 
		Ax^{3}+Bx^{2}+Cx+D \\ 
		0%
	\end{array}
	\right.
	\begin{array}{c}
		x\leq d_{1} \\ 
		d_{1}< x< d_{2} \\ 
		d_{2}\leq x%
	\end{array}
	\label{zerofunction}
\end{equation*}
with $A=-2\left/\left(d_{1}-d_{2}\right)^{3}\right.,$ $B=3\left(d_{1}+d_{2}%
\right)\left/\left(d_{1}-d_{2}\right)^{3}\right.,$ $C=-6d_{1}d_{2}\left/%
\left(d_{1}-d_{2}\right)^{3}\right.$, $D=d_{2}^{2}\left(3d_{1}-d_{2}%
\right)\left/\left(d_{1}-d_{2}\right)^{3}\right.$. 
And the other is
\begin{equation*}
	s\left(x,\epsilon_{\text{s}}\right)=\left \{ 
	\begin{array}{c}
		x \\ 
		\left(1-\epsilon_{\text{s}}\right)+\sqrt{\epsilon_{\text{s}%
			}^{2}-\left(x-x_{2}\right)^{2}} \\ 
		1%
	\end{array}%
	\right.%
	\begin{array}{c}
		0\leq x\leq x_{1} \\ 
		x_{1}< x< x_{2} \\ 
		x_{2}\leq x%
	\end{array}
	\label{sat}
\end{equation*}
with $x_{2}=1+\frac{1}{\tan67.5^{\circ}}\epsilon_{\text{s}}$ and $%
x_{1}=x_{2}-\sin45^{\circ}\epsilon_{\text{s}}.$ Then we design a nominal Lyapunov-like barrier function as \cite{gao2022distributed}
\begin{equation}
	V_{\text{n}}\left(k,x,d_1,d_2,\epsilon,\epsilon_{\text{s}}\right)=\frac{k\sigma\left( x,d_{1},d_{2} \right) }{\left( 1+\epsilon \right) x-d_{1}s\left( \frac{x}{d_{1}},\epsilon _{\text{s}}\right) },
	\label{Vs}
\end{equation}
where $k,\epsilon,\epsilon _{\text{s}}>0$. The function $V_{\text{n}}$ is always continuous and smooth, and has the following properties: (1) $\partial V_{\text{n}} \left/\partial x \right.\leq0$; (2) if $x>d_2$, then $V_{\text{n}}=0$ and $\partial V_{\text{n}} \left/\partial x \right.=0$; if $V_{\text{n}}=0$, then $x>d_2>d_1$; (3) if $0<x<d_1$, then there exists a sufficiently small $\epsilon>0$ such that $V_{\text{n}}\approx \frac{k}{\epsilon x }\geq \frac{k}{\epsilon d_1}. $
In this work, the Lyapunov-like barrier function is used to represent the repulsive potential field.

\subsubsection{Relative Kinematics under Reference Velocity Command}
When robots are moving inside $\mathcal{T}$, the velocities of neighboring robots are approximately the same. Hence, we divide the velocity command $\mathbf{v}_{\text{c},i}$ into two parts, which are the \emph{reference velocity command} $\mathbf{v}^\ast$ for all robots and the \emph{relative velocity command} $\tilde{\mathbf{v}}_{\text{c},i}$ only for the $i$th robot, namely 
\begin{align}
	\mathbf{v}_{\text{c},i}=\tilde{\mathbf{v}}_{\text{c},i}+\mathbf{v}^\ast. \label{vcipartion}
\end{align}
Then, according to the single integrator model \eqref{Single}, the position of the $i$th robot is given by 
$
	\mathbf{p}_{i}=\tilde{\mathbf{p}}_{i}+\mathbf{p}^\ast,
$
where $\tilde{\mathbf{p}}_{i}=\left[\tilde{x}_i\ \tilde{y}_i\right]^{\text{T}}$ is the \emph{relative swarm position} of the $i$th robot, and $\mathbf{p}^\ast \in \mathbb{R}^2$ is the \emph{reference position} of all robots. The reference position $\mathbf{p}^\ast$ can also be considered as the origin of the local coordinate system of the robotic swarm. 
Besides, we have $ \dot{{\mathbf{p}}}^\ast =\mathbf{v}^\ast$ and
\begin{align}
	\dot{\tilde{\mathbf{p}}}_{i} & =\tilde{\mathbf{v}}_{\text{c},i}.  \label{Single1}
\end{align}

\subsection{Distributed Vector Field Controller Design within a Trapezoid Virtual Tube}
\subsubsection{Line Approaching Term and Uniform Flow Field}
For the $i$th robot to approach the finishing line $\left[ {{\mathbf{p}}_{\text{r,}1},{\mathbf{p}}_{\text{h,}1}}\right]$ of $\mathcal{T}$, a \emph{line approaching term} is designed as 
\begin{equation}
	\mathbf{u}_{1,i}=
	\begin{cases}
		v\mathbf{n}_{\text{c}} & \text{if }\mathbf{p}_i \in \mathcal{M}_\mathcal{T} \\
		v\mathbf{t}_{\text{h}} & \text{if }\mathbf{p}_i \in \mathcal{L}_\mathcal{T} \\
		v\mathbf{t}_{\text{r}} & \text{if }\mathbf{p}_i \in \mathcal{R}_\mathcal{T} 
	\end{cases}, \label{u1i}
\end{equation} 
where $v>0$ is the forward speed. 
Define two vectors as $\mathbf{e}_{1}=\left[1\ 0\right]^{\text{T}}$ and $\mathbf{e}_{2}=\left[0\ 1\right]^{\text{T}}$. Then, a uniform flow field for the $i$th robot is defined as 
\begin{align}
	V_{\text{f},i}=U_i\tilde{\mathbf{p}}_{i}^{\text{T}}\left[
	\begin{matrix}
		\cos\alpha_i \\
		\sin\alpha_i
	\end{matrix}\right]
	+c_{i},
\end{align}
where $c_{i}>0$, $U_i=\left \Vert  \mathbf{u}_{1,i}-\mathbf{v}^\ast \right \Vert$ and
$
	\alpha_i=\text{atan2}\left(-\mathbf{e}_2^{\text{T}}\left(\mathbf{u}_{1,i}-\mathbf{v}^\ast\right),-\mathbf{e}_1^{\text{T}}\left(\mathbf{u}_{1,i}-\mathbf{v}^\ast\right)\right).
$
The function $\text{atan2}\left(y,x\right)$ is the four-quadrant arc tangent function. We have $-\left(\partial V_{\text{f},i}/\partial \tilde{\mathbf{p}}_i\right)^{\text{T}}=\mathbf{u}_{1,i}-\mathbf{v}^\ast$.
For the convergence analysis, $c_{i}$ should be chosen carefully such that $V_{\text{f},i}>0$ for any $\mathbf{p}_i \in \mathcal{T}$.

\subsubsection{Barrier Function for Avoiding Collision among Robots}
Define the relative position between the $i$th and $j$th robots as
$
\tilde{\mathbf{p}}_{\text{m},ij} = \mathbf{p}_{i}-{{\mathbf{p}}_{j}}.
$
It should be noted that ${\tilde{\mathbf{p}}}_{i}$ and $\tilde{\mathbf{p}}_{\text{m},ij}$ are not related.
Then, according to the nominal barrier function \eqref{Vs}, the barrier function used to ensure collision avoidance among robots is defined as 
\begin{equation}
	V_{\text{m},ij}=V_{\text{n}}\left(k_2,\left \Vert \tilde{\mathbf{{p}}}{_{\text{m,}ij}}\right \Vert,2r_{\text{s}},r_{\text{s}}+r_{\text{a}},\epsilon_{\text{m}},\epsilon_{\text{s}}\right),
	\label{Vmij}
\end{equation}
where $k_2,\epsilon _{\text{m}},\epsilon _{\text{s}}>0$. The aim of our controller is to minimize $V_{\text{m},ij}$. This ensures that $
\left \Vert \tilde{\mathbf{{p}}}{_{\text{m,}ij}}\right \Vert >2r_{\text{s}}$ and the $i$th robot will have no conflict with the $j$th robot. 

\subsubsection{Barrier Functions for Keeping within Trapezoid Virtual Tube}

\begin{figure}[!t]
	\begin{centering}
		\includegraphics[scale=1]{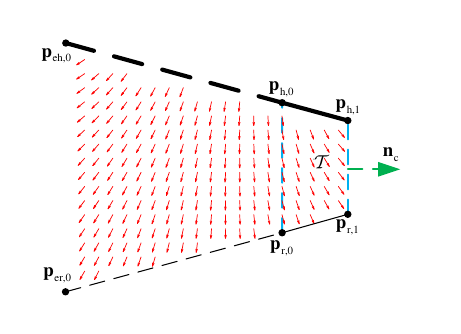}
		\par \end{centering}
	\caption{Repulsive vector field of the extended tube boundary $\left[\mathbf{p}_{\text{eh},0},{\mathbf{p}}_{\text{h},1}\right]$.}
	\label{LRVF}
\end{figure}

Panel methods are widely used in aerodynamic calculations to obtain the solution to the potential flow problem around arbitrarily shaped bodies \cite{kim1992real}. In this work, the single panel method is used to represent the repulsive potential field of the virtual tube boundary \cite{gao2022distributed}. Specifically, given a line segment $\left[\mathbf{a},\mathbf{b}\right]$, the induced repulsive potential function at any point $\mathbf{p}$ by this ``panel" is expressed as 
\begin{align}
	\phi\left(\mathbf{p},\mathbf{a},\mathbf{b},r\right) =\int\limits_{0}^{\left \Vert\mathbf{b}-\mathbf{a}\right \Vert}\ln \left(\left \Vert \mathbf{p}-\left(\mathbf{a}+x\frac{\mathbf{b}-\mathbf{a}}{\left \Vert\mathbf{b}-\mathbf{a}\right \Vert}\right)\right \Vert- r\right)\text{d}x, \label{SinglePanel} 
\end{align}
where $r\geq0$ is the threshold distance. 

To ensure system convergence, we define two extended tube boundaries, namely $\left[\mathbf{p}_{\text{eh},0},{\mathbf{p}}_{\text{h},1}\right]$ and $\left[\mathbf{p}_{\text{er},0},{\mathbf{p}}_{\text{r},1}\right]$. As shown in Figure \ref{LRVF}, these extended tube boundaries satisfy 
	$\left[\mathbf{p}_{\text{h},0},{\mathbf{p}}_{\text{h},1}\right]\subseteq \left[\mathbf{p}_{\text{eh},0},{\mathbf{p}}_{\text{h},1}\right]$, $\left[\mathbf{p}_{\text{r},0},{\mathbf{p}}_{\text{r},1}\right] \subseteq \left[\mathbf{p}_{\text{er},0},{\mathbf{p}}_{\text{r},1}\right].$
Then, two barrier functions used to ensure the $i$th robot keeping within the trapezoid virtual tube are defined as
\begin{align}
	V_{\text{th},i}&=k_{3}\phi\left(\mathbf{p}_i,\mathbf{p}_{\text{eh},0},\mathbf{p}_{\text{h},1},r_\text{s}\right), \label{Vtli} \\ 	V_{\text{tr},i}&=k_{3}\phi\left(\mathbf{p}_i,\mathbf{p}_{\text{er},0},\mathbf{p}_{\text{r},1},r_\text{s}\right), \label{Vtri}
\end{align}
where $k_3>0$. The points $\mathbf{p}_{\text{eh},0}$ and $\mathbf{p}_{\text{er},0}$ are subject to the following conditions
\begin{equation}
	\left\{
	\begin{aligned}
		-\frac{\partial V_{\text{th},i}}{\partial \mathbf{p}_i }\mathbf{n}_{\text{c}}&\geq0 \text{ if } \mathbf{p}_i \in \mathcal{T}\\
		-\frac{\partial V_{\text{tr},i}}{\partial \mathbf{p}_i }\mathbf{n}_{\text{c}}&\geq0 \text{ if } \mathbf{p}_i \in \mathcal{T}  
	\end{aligned}
	\right..
	\label{DirL2}
\end{equation}
As shown in Figure \ref{LRVF}, the conditions in \eqref{DirL2} indicate that within $\mathcal{T}$, the angles between $\mathbf{n}_{\text{c}}$ and the negative gradient directions of $V_{\text{th},i}$ and $V_{\text{tr},i}$ should remain less than $90^{\circ}$. If the lengths of $\left[\mathbf{p}_{\text{eh},0},{\mathbf{p}}_{\text{h},1}\right]$ and $\left[\mathbf{p}_{\text{er},0},{\mathbf{p}}_{\text{r},1}\right]$ are sufficiently large, the constraints \eqref{DirL2} are guaranteed to be satisfied. The aim of the controller design is to minimize $V_{\text{th},i}$ and $V_{\text{tr},i}$. This ensures that $d_{\mathcal{T}\text{h}}\left(\mathbf{p}_i\right) >r_\text{s}$ and $d_{\mathcal{T}\text{r}}\left(\mathbf{p}_i\right)>r_\text{s}$.

\textbf{Remark 3}. The conditions in \eqref{DirL2} are necessary for the safety and convergence proofs in the following subsections. Take the left boundary $\left[ \mathbf{p}_{\text{h},0},{\mathbf{p}}_{\text{h},1} \right]$ as an example. If we have $\mathbf{n}_{\text{c}}^{\text{T}}\mathbf{n}_\text{h}<0$, the extended tube boundary must be longer than original tube boundary, namely $\left\Vert\mathbf{p}_{\text{eh},0}-\mathbf{p}_{\text{h},1}\right\Vert>\left\Vert\mathbf{p}_{\text{h},0}-\mathbf{p}_{\text{h},1}\right\Vert$ is satisfied. If we have $\mathbf{n}_{\text{c}}^{\text{T}}\mathbf{n}_\text{h}\geq0$, the condition in \eqref{DirL2} is already satisfied for the original tube boundary, namely $\mathbf{p}_{\text{eh},0}=\mathbf{p}_{\text{h},0}$ is feasible. A similar analysis can also be performed on the right boundary $\left[ \mathbf{p}_{\text{r},0},{\mathbf{p}}_{\text{r},1} \right]$. In the passing-through control problem, the circumstance of $\mathbf{n}_{\text{c}}^{\text{T}}\mathbf{n}_\text{h}<0,\mathbf{n}_{\text{c}}^{\text{T}}\mathbf{n}_\text{r}<0$ is common.


\subsubsection{Distributed Vector Field Controller Design}

A swarm controller for the $i$th robot is formulated as 
\begin{equation}
	\mathbf{v}_{\text{c},i}=\mathbf{u}_{1,i}+\text{sat}\left(\mathbf{u}_{2,i}+\mathbf{u}_{3,i}+\mathbf{u}_{4,i},{v_{\text{max}}^{\prime}}\right), \label{VectorField}
\end{equation}
where the line approaching term $\mathbf{u}_{1,i}$ is defined in \eqref{u1i}, and the constant ${v_{\text{max}}^{\prime}}>0$ represents the maximum norm of the repulsive vector field. The subcommands $\mathbf{u}_{2,i}$, $\mathbf{u}_{3,i}$, $\mathbf{u}_{4,i}$ are called \emph{robot avoidance term}, \emph{left boundary avoidance term} and \emph{right boundary avoidance term}, respectively. These subcommands are written as
\begin{align*}
	\mathbf{u}_{2,i}&=\sum_{j\in \mathcal{N}_{\text{m},i}}
	\underbrace{-\frac{\partial V_{\text{m},ij}}{\partial \left \Vert \tilde{\mathbf{{p}}}{_{\text{m,}ij}}\right \Vert }\frac{1}{\left \Vert \tilde{\mathbf{{p}}}{_{\text{m,}ij}}\right \Vert }}_{b_{ij}}
	\tilde{\mathbf{p}}_{\text{m},ij}=\sum_{j\in \mathcal{N}_{\text{m},i}}
	b_{ij}\tilde{\mathbf{p}}_{\text{m},ij},\\
	\mathbf{u}_{3,i}&=-\left(\frac{\partial V_{\text{th},i}}{\partial \mathbf{p}_i }\right)^{\text{T}},\\
	\mathbf{u}_{4,i}&=-\left(\frac{\partial V_{\text{tr},i}}{\partial \mathbf{p}_i }\right)^{\text{T}}.
\end{align*}
When all robots can get the information of the trapezoid virtual tube and the relative position information of other neighboring robots, the swarm controller \eqref{VectorField} is fully distributed and can work autonomously without wireless communication or IDs of neighboring robots.
Besides, to satisfy the constraints \eqref{vminvmax}, it is required that $\left \Vert \mathbf{u}_{1,i} \right \Vert +{v_{\text{max}}^{\prime}}\leq v_{\text{max}}$ and $\left \Vert \mathbf{u}_{1,i} \right \Vert -{v_{\text{max}}^{\prime}}\geq v_{\text{min}} $. Then, according to \eqref{u1i}, two necessary constraints are written as
\begin{equation}
	\left\{
	\begin{aligned}
		&v+{v_{\text{max}}^{\prime}} \leq v_{\text{max}}\\
		&v-{v_{\text{max}}^{\prime}} \geq v_{\text{min}} 
	\end{aligned}
	\right..
	\label{Constraint1}
\end{equation}

\subsection{Convergence Analysis}
\begin{figure}[!t]
	\centering
	\includegraphics[scale=1.1]{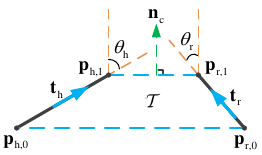}
	\caption{The introduction of $\theta_\text{h}$ and $\theta_\text{r}$.}
	\label{thetalr}
\end{figure}

Before presenting the convergence analysis, some necessary constraints for the trapezoid virtual tube are proposed as
\begin{equation}
	\begin{cases}
		v\left\Vert \mathbf{n}_{\text{c}} -\mathbf{t}_{\text{h}} \right \Vert \leq {v_{\text{max}}^{\prime}}  & \text{if } \mathcal{L}_\mathcal{T}\neq \varnothing\\
		v\left\Vert \mathbf{n}_{\text{c}} -\mathbf{t}_{\text{r}} \right \Vert \leq {v_{\text{max}}^{\prime}} & \text{if } \mathcal{R}_\mathcal{T}\neq \varnothing
	\end{cases}
	\label{Constraint2}.
\end{equation}
The constraints in \eqref{Constraint2} show the relationship between the trapezoid virtual tube and the speed constraints. As shown in Figure \ref{thetalr}, these constraints imply that if $ \mathcal{L}_\mathcal{T}\neq \varnothing$ and $\mathcal{R}_\mathcal{T}\neq \varnothing$ are both satisfied, 
\begin{equation*}
\max\left(\theta_\text{h},\theta_\text{r}\right) \leq \arccos\left(1-v_{\text{max}}^{\prime2}/2v^2\right)
\end{equation*}
should be satisfied,
where $\theta_\text{h}=\arccos\left(\mathbf{t}_{\text{h}}^{\text{T}}\mathbf{n}_{\text{c}}\right)$ and $\theta_\text{r}=\arccos\left(\mathbf{t}_{\text{r}}^{\text{T}}\mathbf{n}_{\text{c}}\right)$. Then we have the following lemma.

\textbf{Lemma 1}. Under \textit{Assumptions 1-4}, suppose that (i) the velocity command is designed as \eqref{VectorField}; (ii) the constraints \eqref{DirL2}, \eqref{Constraint1}, \eqref{Constraint2} are satisfied. Then, there exist sufficiently small $\epsilon _{\text{m}},\epsilon _{\text{s}}>0$
in \eqref{Vmij} such that $\left \Vert \tilde{\mathbf{{p}}}{_{\text{m,}ij}}\left( t\right)\right\Vert >2r_{\text{s}},$ $d_{\mathcal{T}\text{h}}\left(\mathbf{p}_i\left( t\right)\right) >r_\text{s}$, $d_{\mathcal{T}\text{r}}\left(\mathbf{p}_i\left( t\right)\right)>r_\text{s}$, $t\in \lbrack 0,\infty )$ for all ${{\mathbf{p}}_{i}(0)}$, $i,j=1,\cdots ,M,i\neq j.$

\emph{Proof}. See \emph{Appendix A}. $\square$

With \emph{Lemma 1} in hand, a theorem is stated as follows. 

\textbf{Theorem 1}. Under \textit{Assumptions 1-4}, suppose that
(i) the velocity command is designed as \eqref{VectorField};  (ii) the constraints \eqref{DirL2}, \eqref{Constraint1}, \eqref{Constraint2} are satisfied. Then, given ${\epsilon }_{\text{0}}>0$, there
exist sufficiently small $\epsilon _{\text{m}},\epsilon _{\text{s}}>0$
in \eqref{Vmij} and $t_{1}>0$ such that all
robots can satisfy \eqref{TraCondition} at $t\geq t_{1},$ meanwhile ensuring  
$\mathcal{S}_{i}\left(t\right)\cap  \mathcal{S}_{j}\left(t\right)=\varnothing $, $\mathcal{S}_{i}\left(t\right)\cap \partial\mathcal{T}=\varnothing$,  $t\in \lbrack 0,t_1 \rbrack$ for all ${{\mathbf{p}}_{i}(0)}$, $i,j=1,\cdots ,M,i\neq j$.

\emph{Proof}. See \emph{Appendix B}. $\square$

\subsection{Modified Vector Field Controller Design}
Two noticeable limitations exist in the application of the controller \eqref{VectorField}.
\begin{itemize}
	\item The first limitation is that $\mathbf{u}_{1,i}$ is not continuous or smooth when $\mathbf{p}_{i}$ is switched between $\mathcal{M}_\mathcal{T}$ and $\mathcal{L}_\mathcal{T}$ or between $\mathcal{M}_\mathcal{T}$ and $\mathcal{R}_\mathcal{T}$.
	\item The second limitation is that it is challenging to obtain the values of $\mathbf{p}_{\text{eh},0}$ and $\mathbf{p}_{\text{er},0}$ in \eqref{Vtli}, \eqref{Vtri}. Furthermore, due to the existence of integrals in \eqref{SinglePanel}, the mathematical expressions of $\mathbf{u}_{3,i}$ and $\mathbf{u}_{4,i}$ are complex and not convenient for implementation.
\end{itemize}

To make the controller continuous and smooth, a modified line approaching term $\mathbf{u}_{1,i}^{\prime\prime}$ is designed as
\begin{equation}
	\mathbf{u}_{1,i}^{\prime\prime}=
	\begin{cases}
		v\mathbf{n}_{\text{c}} & \text{if }\mathbf{p}_i \in \mathcal{M}_\mathcal{T} \\
		v\frac{\sigma\left( d_{\mathcal{T}\text{h}}\left(\mathbf{p}_i\right),r_\text{a},k_\text{t} r_\text{a}  \right)\left(\mathbf{t}_{\text{h}}-\mathbf{n}_{\text{c}}\right)+\mathbf{n}_{\text{c}}}{\left\Vert \sigma\left( d_{\mathcal{T}\text{h}}\left(\mathbf{p}_i\right) ,r_\text{a},k_\text{t} r_\text{a} \right)\left(\mathbf{t}_{\text{h}}-\mathbf{n}_{\text{c}}\right)+\mathbf{n}_{\text{c}} \right\Vert} & \text{if }\mathbf{p}_i \in \mathcal{L}_\mathcal{T} \\
		v\frac{\sigma\left( d_{\mathcal{T}\text{r}}\left(\mathbf{p}_i\right) ,r_\text{a},k_\text{t} r_\text{a} \right)\left(\mathbf{t}_{\text{r}}-\mathbf{n}_{\text{c}}\right)+\mathbf{n}_{\text{c}}}{\left\Vert \sigma\left( d_{\mathcal{T}\text{r}}\left(\mathbf{p}_i\right) ,r_\text{a},k_\text{t} r_\text{a} \right)\left(\mathbf{t}_{\text{r}}-\mathbf{n}_{\text{c}}\right)+\mathbf{n}_{\text{c}} \right\Vert} & \text{if }\mathbf{p}_i \in \mathcal{R}_\mathcal{T} 
	\end{cases},
\end{equation}
where $d_{\mathcal{T}\text{h}}\left(\mathbf{p}_i\right) $, $d_{\mathcal{T}\text{r}}\left(\mathbf{p}_i\right)$ are defined in \eqref{dtl}, \eqref{dtr}. Compared to $\mathbf{u}_{1,i}$ defined in \eqref{u1i}, smooth transition processes are added to the modified subcommand $\mathbf{u}_{1,i}^{\prime\prime}$. Two non-potential control terms are proposed to approximate the behavior of $\mathbf{u}_{3,i}$ and $\mathbf{u}_{4,i}$ as stated in \cite{gao2022distributed}. Two barrier functions are defined as
\begin{align}
	V_{\text{th},i}^{\prime}&=V_{\text{n}}\left(k_3,d_{\mathcal{T}\text{h}}\left(\mathbf{p}_i\right),r_{\text{s}},r_{\text{a}},\epsilon_{\text{t}},\epsilon_{\text{s}}\right), \label{Vtli1}\\
	V_{\text{tr},i}^{\prime}&=V_{\text{n}}\left(k_3,d_{\mathcal{T}\text{r}}\left(\mathbf{p}_i\right),r_{\text{s}},r_{\text{a}},\epsilon_{\text{t}},\epsilon_{\text{s}}\right), \label{Vtri1}
\end{align}
where $k_3,\epsilon _{\text{t}}>0$. The aim of the controller design is to minimize $V_{\text{th},i}^{\prime}$ and $V_{\text{tr},i}^{\prime}$. This ensures that $d_{\mathcal{T}\text{h}}\left(\mathbf{p}_i\right) >r_\text{s}$ and $d_{\mathcal{T}\text{r}}\left(\mathbf{p}_i\right)>r_\text{s}$.

\begin{figure}[!t]
	\begin{centering}
		\includegraphics[scale=0.35]{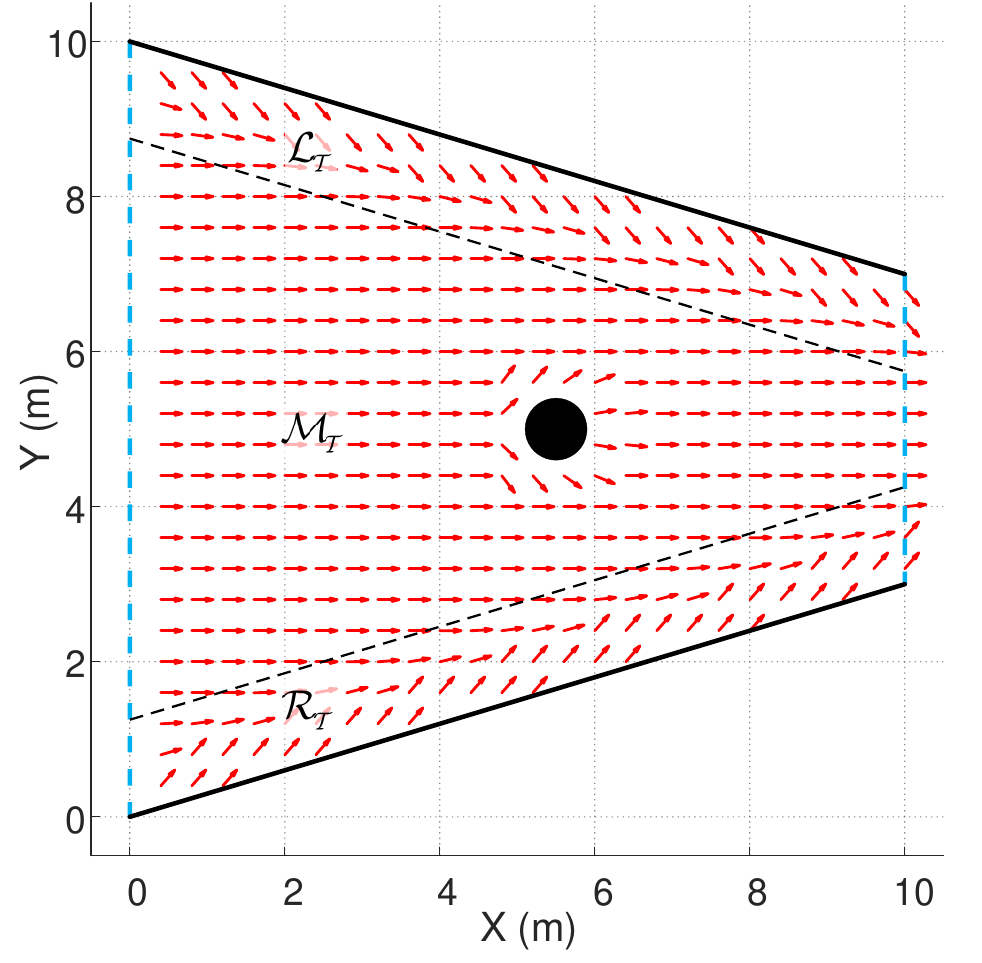}
		\par \end{centering}
	\caption{Vector field with the swarm controller \eqref{ModifiedVectorField}. The black circle represents another robot. }
	\label{VFplot}
\end{figure}

Define $\mathbf{p}$ as the collection $\left( \mathbf{p}_{1},\cdots ,\mathbf{p}_{M}\right) $. The modified distributed swarm controller is designed as 
\begin{align}
	\mathbf{v}_{\text{c},i}\!=\!\mathbf{v}\left( \mathcal{T},\mathbf{p}_{i},
	\mathbf{p}\right) 
	\!=\!\mathbf{u}_{1,i}^{\prime\prime}\!+\!\text{sat}\left(\mathbf{u}_{2,i}\!+\!\mathbf{u}_{3,i}^{\prime}\!+\!\mathbf{u}_{4,i}^{\prime},{v_{\text{max}}^{\prime}}\right). \label{ModifiedVectorField}
\end{align}
The modified subcommands $\mathbf{u}_{3,i}^{\prime}$ and $\mathbf{u}_{4,i}^{\prime}$ are written as 
\begin{align*}
	\mathbf{u}_{3,i}^{\prime}&=-\frac{\partial V_{\text{th},i}^{\prime}}{\partial d_{\mathcal{T}\text{h}}\left(\mathbf{p}_i\right)  }\mathbf{P}_{\text{t}}\mathbf{n}_\text{h}, \\
	\mathbf{u}_{4,i}^{\prime}&=-\frac{\partial V_{\text{tr},i}^{\prime}}{\partial d_{\mathcal{T}\text{r}}\left(\mathbf{p}_i\right)  }\mathbf{P}_{\text{t}}\mathbf{n}_\text{r},
\end{align*}
where $\mathbf{P}_{\text{t}}=\mathbf{P}_{\text{t}}^{\text{T}}=\mathbf{n}_{\text{c}}\mathbf{n}_{\text{c}}^{\text{T}}$ maps a vector in the direction of $\mathbf{n}_{\text{c}}.$ The modified controller \eqref{ModifiedVectorField} is always continuous and smooth. Figure \ref{VFplot} shows the vector field of the modified controller \eqref{ModifiedVectorField}. In Figure \ref{VFplot}, the black circle represents another robot.

In the modified controller \eqref{ModifiedVectorField}, there are twelve parameters that can be tuned, including $r_\text{s}$, $r_\text{a}$, $v_\text{min}$, $v_\text{max}$, $v_\text{max}^{\prime}$, $k_\text{t}$, $v$, $k_2$, $\epsilon_\text{m}$, $\epsilon_\text{s}$, $k_3$, $\epsilon_\text{t}$. The safety radius $r_\text{s}$ chosen should be a little larger than the physical radius of the robot. The avoidance radius $r_\text{a}$ chosen should be smaller than the detection radius of the robot. The minimum and the maximum permitted speeds $v_\text{min}$, $v_\text{max}$ should satisfy the physical constraints of the robot. The parameters $v_\text{max}^{\prime}$ and $v$ should satisfy the constraints in \eqref{Constraint1}. A common set of critical values are $v_\text{max}^{\prime}=\frac{1}{2}\left(v_\text{max}-v_\text{min}\right)$, $v=\frac{1}{2}\left(v_\text{max}+v_\text{min}\right)$. The parameter $k_\text{t}$ chosen can be a bit larger than 1. According to our simulations and experiments, $k_\text{t}=1.5$ is usually an acceptable choice. The rest of the parameters $k_2$, $\epsilon_\text{m}$, $\epsilon_\text{s}$, $k_3$, $\epsilon_\text{t}$ are related to the barrier functions. To ensure convergence, $\epsilon_\text{m}$, $\epsilon_\text{t}$ should be set as tiny values, such as $\epsilon_\text{m}=\epsilon_\text{t}=10^{-6}$. The control gains $k_2$, $k_3$ and the parameter $\epsilon_\text{s}$ have a relatively small impact on the control effect. And $k_2=k_3=1$, $\epsilon_\text{s}=10^{-6}$ are suitable for most scenarios.

\textbf{Remark 4}. The modification from $\mathbf{u}_{1,i}$, $\mathbf{u}_{3,i}$, $\mathbf{u}_{4,i}$ to $\mathbf{u}_{1,i}^{\prime\prime}$, $\mathbf{u}_{3,i}^{\prime}$, $\mathbf{u}_{4,i}^{\prime}$ has no negative effect on the convergence of the robotic swarm system. Firstly, the subcommands $\mathbf{u}_{1,i}$, $\mathbf{u}_{1,i}^{\prime\prime}$ have the same variation range of the vector direction, which guarantees that the proofs of \textit{Lemma 1} and \textit{Theorem 1} are not violated according to the constraints \eqref{Constraint2}. Then, we will explain why $\mathbf{u}_{3,i}^{\prime}$ can approximate $\mathbf{u}_{3,i}$. Define a vector as $\mathbf{u}_{\text{he},i}=-\left(\partial V_{\text{th},i}^{\prime}/\partial d_{\mathcal{T}\text{h}}\left(\mathbf{p}_i\right)\right) \mathbf{n}_\text{h}$, which is the negative gradient of $V_{\text{th},i}^{\prime}$ and perpendicular to the line segment $\left[\mathbf{p}_{\text{h},0},{\mathbf{p}}_{\text{h},1}\right]$. And $	\mathbf{u}_{3,i}^{\prime}=\mathbf{P}_{\text{t}}\mathbf{u}_{\text{he},i}$ is a non-potential term and always orthogonal to $\mathbf{n}_{\text{c}}$. To ensure system convergence, it is not feasible to directly apply $\mathbf{u}_{\text{he},i}$ in the modified controller \eqref{ModifiedVectorField}, which may make $\left(\partial V_{\text{th},i}/\partial \mathbf{p}_i+\partial V_{\text{tr},i}/\partial \mathbf{p}_i\right)\mathbf{v}^\ast\leq0$ untenable. Thus, it is necessary to adopt the single panel method in this work. The adoption of $\mathbf{p}_{\text{eh},0}, \mathbf{p}_{\text{er},0}$ instead of $\mathbf{p}_{\text{h},0}, \mathbf{p}_{\text{r},0}$ in \eqref{Vtli}, \eqref{Vtri} is for the same reason. Assume that we have $\left[\mathbf{p}_{\text{h},0},{\mathbf{p}}_{\text{h},1}\right]\subseteq \left[\mathbf{p}_{\text{eh},0},{\mathbf{p}}_{\text{eh},1}\right]$. If $\left[\mathbf{p}_{\text{eh},0},{\mathbf{p}}_{\text{eh},1}\right]$ is long enough, the changes in the orientation of $\mathbf{u}_{3,i}$ within the trapezoid virtual tube may be insignificant. Consequently, it is possible to choose appropriate points $\mathbf{p}_{\text{eh},0}$ and $\mathbf{p}_{\text{eh},1}$ so that the orientations of $\mathbf{u}_{3,i}$ are nearly perpendicular to $\mathbf{n}_{\text{c}}$. This explains the approximation of $\mathbf{u}_{3,i}^{\prime}$ to $\mathbf{u}_{3,i}$. Similar analyses can be made to explain why $\mathbf{u}_{4,i}^{\prime}$ can approximate $\mathbf{u}_{4,i}$. Therefore, the validity of \emph{Lemma 1} and \emph{Theorem 1} is maintained even if we substitute the condition (i) with the velocity command designed as \eqref{ModifiedVectorField}.

\section{Switching Logic for Obstacle Avoidance}
In the last section, the swarm controller \eqref{ModifiedVectorField} is designed for the trapezoid virtual tube under speed constraints. However, the main issue is that no obstacle is considered. For obstacle avoidance inside $\mathcal{T}$, a common idea is to introduce a potential-based obstacle avoidance term into the swarm controller. This method is not feasible in this paper. Although obstacles are static, the robots are actually faced with the dynamic obstacle avoidance problem due to the existence of the reference velocity command $\mathbf{v}^\ast$. The detailed description is given in \emph{Appendix C}. Hence, in this section, we partition a trapezoid virtual tube that contains obstacles into multiple sub trapezoid virtual tubes without obstacles. The corresponding switching logic is also proposed. 

\subsection{Partition of a Trapezoid Virtual Tube}

\begin{figure}[!t]
	\centerline{\includegraphics[width=\columnwidth]{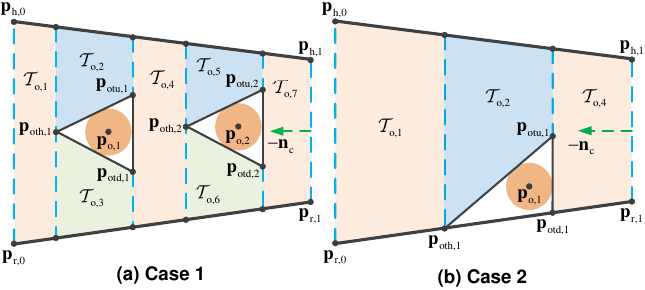}}
	\caption{Partition of the trapezoid virtual tube for static obstacle avoidance. In plot (b), there exists $\mathcal{T}_{\text{o},3}=\varnothing$.}
	\label{ObstacleTube}
\end{figure}

According to \eqref{obstacle}, there are $P$ obstacles inside the trapezoid virtual tube $\mathcal{T}$. For simplicity, the obstacle that is the farthest from the finishing line $\left[ {{\mathbf{p}}_{\text{r,}1},{\mathbf{p}}_{\text{h,}1}}\right]$ is labeled as the 1st obstacle. Correspondingly, the obstacle closest to the finishing line is labeled as the $P$th obstacle. Then, the \emph{extended occupied area} of the $k$th obstacle is designed as 
\begin{equation}
	\mathcal{O}_{k}^{\prime}=\left \{ \mathbf{x}\in {{\mathbb{R}}^{2}}: \left \Vert \mathbf{x}-\mathbf{p}_{\text{o},k}\right \Vert \leq r_{\text{o},k}+r_{\text{s}} \right \}. \label{ExtObstacle}
\end{equation}
Besides, a \emph{circumscribed isosceles triangle} $\mathcal{I}_k$ is proposed for the $k$th obstacle, whose vertex is $\mathbf{p}_{\text{oth},k}$, base points are $\mathbf{p}_{\text{otu},k}$ and $\mathbf{p}_{\text{otd},k}$. It is required that 
$\frac{\mathbf{p}_{\text{o},k}-\mathbf{p}_{\text{oth},k}}{\left \Vert \mathbf{p}_{\text{o},k}-\mathbf{p}_{\text{oth},k} \right \Vert}=\mathbf{n}_\text{c}$ and $
\mathbf{n}_\text{c}^{\text{T}}\left(\mathbf{p}_{\text{otu},k}-\mathbf{p}_{\text{otd},k}\right)=0$. For the $k$th obstacle, its $\mathcal{I}_k$ is not unique. The shape and the size of $\mathcal{I}_k$ are determined by $\left\Vert \mathbf{p}_{\text{oth},k}-\mathbf{p}_{\text{o},k} \right\Vert$, which is adjustable and should satisfy the constraint \eqref{pothk} described in the following. As shown in Figure \ref{ObstacleTube}, we have $\mathcal{O}_{k} \subset \mathcal{O}_{k}^{\prime} \subset \mathcal{I}_{k}$, $k=1,\cdots,P$.

With circumscribed isosceles triangles $\mathcal{I}_k,k=1,\cdots,P$ available, the trapezoid virtual tube $\mathcal{T}$ is divided into $\left(3P+1\right)$ sub trapezoid virtual tubes without obstacles, namely $\mathcal{T}_{\text{o},k},k=1,\cdots,3P+1$. As shown in Figure \ref{ObstacleTube}, we have
\begin{align}
	&\left(\cup_{k=1}^{3P+1}\mathcal{T}_{\text{o},k}\right)\cup \left(\cup_{k=1}^{P}\mathcal{I}_k \right)=\mathcal{T}, \label{FourSub1}\\
	&\left(\cup_{k=1}^{3P+1}\mathcal{T}_{\text{o},k}\right)\cap \left(\cup_{k=1}^{P}\mathcal{I}_k \right)=\varnothing. \label{FourSub2}
\end{align}

\textbf{Remark 5}. In real practice, \emph{Assumption 3} may not be satisfied for all sub trapezoid virtual tubes, which means that these sub trapezoid virtual tubes are too narrow or too short to pass through. For example, as shown in Figure \ref{ObstacleTube}(b), the sub trapezoid virtual tube $\mathcal{T}_{\text{o},3}$ is too narrow. Under this circumstance, we let $\mathcal{T}_{\text{o},3}=\varnothing$, and the circumscribed triangle $\mathcal{I}_1$ is modified as $\frac{\mathbf{p}_{\text{otd},1}-\mathbf{p}_{\text{oth},1}}{\left \Vert \mathbf{p}_{\text{otd},1}-\mathbf{p}_{\text{oth},1} \right \Vert}=\mathbf{t}_\text{r},\mathbf{n}_\text{c}^{\text{T}}\left(\mathbf{p}_{\text{otu},k}-\mathbf{p}_{\text{otd},k}\right)=0$. It should be noted that this triangle is no longer isosceles. With this modification, robots can pass through the trapezoid virtual tube successfully and safely.

\subsection{Switching Logic Design for Obstacle Avoidance}

\begin{figure}[!t]
	\begin{centering}
		\includegraphics[scale=0.35]{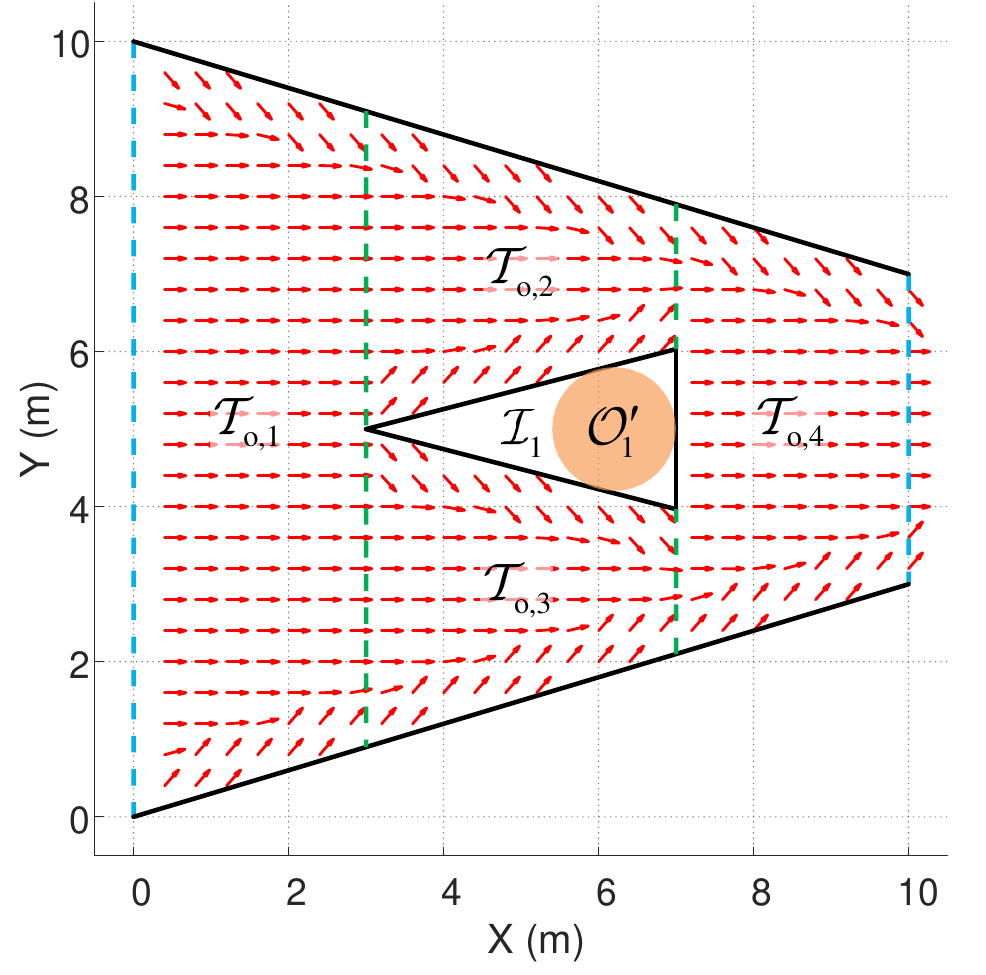}
		\par \end{centering}
	\caption{Vector field of a trapezoid virtual tube containing a static obstacle with the swarm controller \eqref{ObstacleController}.}
	\label{VFObstacle}
\end{figure}

With sub trapezoid virtual tubes $\mathcal{T}_{\text{o},k}$ available, the switching logic for the $i$th robot to avoid collision with obstacles inside $\mathcal{T}$ is designed as  
\begin{align}
	\mathbf{v}_{\text{c},i}=\mathbf{v}\left( \mathcal{T}_{\text{o},k},\mathbf{p}_{i},
	\mathbf{p}\right) \text{ if } \mathbf{p}_{i} \in \mathcal{T}_{\text{o},k},
	\label{ObstacleController}
\end{align}
where $k=1,\cdots,3P+1$. In a given scenario, a robot is moving inside a trapezoid virtual tube while a static obstacle is positioned inside the virtual tube. In Figure \ref{VFObstacle}, the vector field of the trapezoid virtual tube with the swarm controller \eqref{ObstacleController} is depicted.

Similar to \emph{Assumption 3}, we have a requirement for the virtual tube planning, which is summarized as the following assumption for all sub trapezoid virtual tubes available.

\textbf{Assumption 3$^\prime$}. All sub trapezoid virtual tubes available $\mathcal{T}_{\text{o},k}\neq\varnothing,k=1,\cdots,3P+1$ are long enough and wide enough to contain at least one robot. Mathematically, if $\mathcal{T}_{\text{o},k}\neq \varnothing$, it is required that $\mathcal{L}_{\mathcal{T}_{\text{o},k}} \cap \mathcal{R}_{\mathcal{T}_{\text{o},k}}=\varnothing$, $\mathcal{M}_{\mathcal{T}_{\text{o},k}}\neq\varnothing$, $ \mathbf{n}_{\text{c}}^{\text{T}}\left({\mathbf{p}}_{\text{r,}1}-{\mathbf{p}}_{\text{r,}0}\right)>2r_\text{a}$, where $\mathbf{n}_{\text{c}},{\mathbf{p}}_{\text{r,}1},{\mathbf{p}}_{\text{r,}0}$ are defined in $\mathcal{T}_{\text{o},k}$, and $k=1,\cdots,3P+1$.

Then, if $\mathcal{T}_{\text{o},k}\neq \varnothing$, some constraints similar to \eqref{Constraint2} are shown as 
\begin{equation}
	\begin{cases}
		v\left\Vert \mathbf{n}_{\text{c}} -\mathbf{t}_{\text{h}} \right \Vert\leq{v_{\text{max}}^{\prime}}  & \text{if } \mathcal{L}_{\mathcal{T}_{\text{o},k}}\neq \varnothing\\
		v\left\Vert \mathbf{n}_{\text{c}} -\mathbf{t}_{\text{r}} \right \Vert\leq{v_{\text{max}}^{\prime}} & \text{if } \mathcal{R}_{\mathcal{T}_{\text{o},k}}\neq \varnothing
	\end{cases}
	\label{Constraint3},
\end{equation}
where $\mathbf{t}_{\text{h}} $ and $\mathbf{t}_{\text{r}}$ are defined in $\mathcal{T}_{\text{o},k}$, and $k=1,\cdots,3P+1$. The constraints \eqref{Constraint3} also limit the shape of the circumscribed triangles. If the circumscribed triangle of the $k$th obstacle is isosceles, we have the following constraint as 
\begin{equation}
	\left\Vert \mathbf{p}_{\text{oth},k}-\mathbf{p}_{\text{o},k} \right\Vert\geq\frac{r_s+r_{\text{o},k}}{\frac{v_{\text{max}}^{\prime}}{v}\sqrt{1-\frac{v_{\text{max}}^{\prime2}}{4v^2}}}. \label{pothk}
\end{equation}
The larger $\left\Vert \mathbf{p}_{\text{oth},k}-\mathbf{p}_{\text{o},k} \right\Vert$, the smoother obstacle avoidance behavior the robot has. With these assumptions and constraints in hand, an important theorem is proposed as follows. 

\textbf{Theorem 2}. Under \textit{Assumptions 1-4, 3$^\prime$}, suppose that
(i) the velocity command is designed as \eqref{ObstacleController}; (ii) the constraints \eqref{Constraint1}, \eqref{Constraint2}, \eqref{Constraint3} are satisfied. Then, given ${\epsilon }_{\text{0}}>0$, there
exist sufficiently small $\epsilon _{\text{m}},\epsilon _{\text{s}}>0$
in \eqref{Vmij}, $\epsilon _{\text{t}}>0$
in \eqref{Vtli1}, \eqref{Vtri1} and $t_{1}>0$ such that all
robots can satisfy \eqref{TraCondition} for $\mathcal{T}$ at $t\geq t_{1},$ meanwhile ensuring  
$\mathcal{S}_{i}\left(t\right)\cap  \mathcal{S}_{j}\left(t\right)=\varnothing $, $\mathcal{S}_{i}\left(t\right)\cap \partial\mathcal{T}=\varnothing$, $\mathcal{S}_{i}\left(t\right)\cap  \mathcal{O}_{k}=\varnothing $, $t\in \lbrack 0,t_1 \rbrack$ for all ${{\mathbf{p}}_{i}(0)}$, $i,j=1,\cdots ,M,i\neq j,k=1,\cdots ,P$.


\emph{Proof}. 
According to \emph{Lemma 1}, robots are able to avoid collisions with one another and keep moving inside their corresponding sub trapezoid virtual tubes. According to the extended occupied area defined in \eqref{ExtObstacle}, $\left\Vert\mathbf{p}_i-\mathbf{p}_{\text{o},k}\right\Vert>r_{\text{o},k}+r_{\text{s}},k=1,\cdots ,P$ is always satisfied, namely the $i$th robot can avoid collision with obstacles. According to \emph{Theorem 1}, all robots can pass through these sub trapezoid virtual tubes. Without loss of generality, we assume that there is only one obstacle inside $\mathcal{T}$ and all sub trapezoid virtual tubes satisfy \emph{Assumption 3}. If a robot is inside $\mathcal{T}_{\text{o},1}$ in the beginning, then it has two different possible 
ways to approach $\left[ \mathbf{p}_{\text{h},1},{\mathbf{p}}_{\text{r},1}\right] $, namely $\mathcal{T}_{\text{o},1}\rightarrow \mathcal{T}_{\text{o},2}\rightarrow \mathcal{T}_{\text{o},4}$ and  $\mathcal{T}_{\text{o},1}\rightarrow \mathcal{T}_{\text{o},3}\rightarrow \mathcal{T}_{\text{o},4}$. It should be noted that there exist $3P$ \emph{singular points} inside $\mathcal{T}$, namely $\mathbf{p}_{\text{oth},k}$, $\mathbf{p}_{\text{otu},k}$, $\mathbf{p}_{\text{otd},k}$, $k=1,\cdots ,P$. When robots are just located at these points, they have no idea how to move. However, in real practice, these singular points only form a zero-measure set, namely the possibility of robots located there is approximately zero.  In a word, robots can pass through $\mathcal{T}$ when there exist obstacles. $\square$

\textbf{Remark 6}. In this section, we focus on the controller design when robots are switching between sub trapezoid virtual tubes. How to plan and generate these sub trapezoid virtual tubes is not our focus. In real practice, the partition of the trapezoid virtual tube can be modeled as an optimization problem, whose objective is usually related to maximizing the passing-through efficiency and minimizing energy consumption under the requirement of the constraint \eqref{pothk} and \emph{Assumption 3$^\prime$}. For example, if the robots are fixed-wing UAVs, less turning causes less energy consumption, so a larger $\left\Vert \mathbf{p}_{\text{oth},k}-\mathbf{p}_{\text{o},k} \right\Vert$ is preferred. However, the larger $\left\Vert \mathbf{p}_{\text{oth},k}-\mathbf{p}_{\text{o},k} \right\Vert$, the more space the circumscribed triangle $\mathcal{I}_k$ occupies, which may make the swarm crowded and cause a low passing-through efficiency. Hence, the optimization program should balance these objectives and produce a compromise result.

\section{Simulations and Experiments}
A video about simulations and experiments is available on \href{http://rfly.buaa.edu.cn}{http://rfly.buaa.edu.cn} and
\href{https://youtu.be/RBnK97e8S4k}{https://youtu.be/RBnK97e8S4k}. 

\subsection{Numerical Simulation}
\begin{figure}[!t]
	\centering
	\includegraphics[width=0.9\columnwidth]{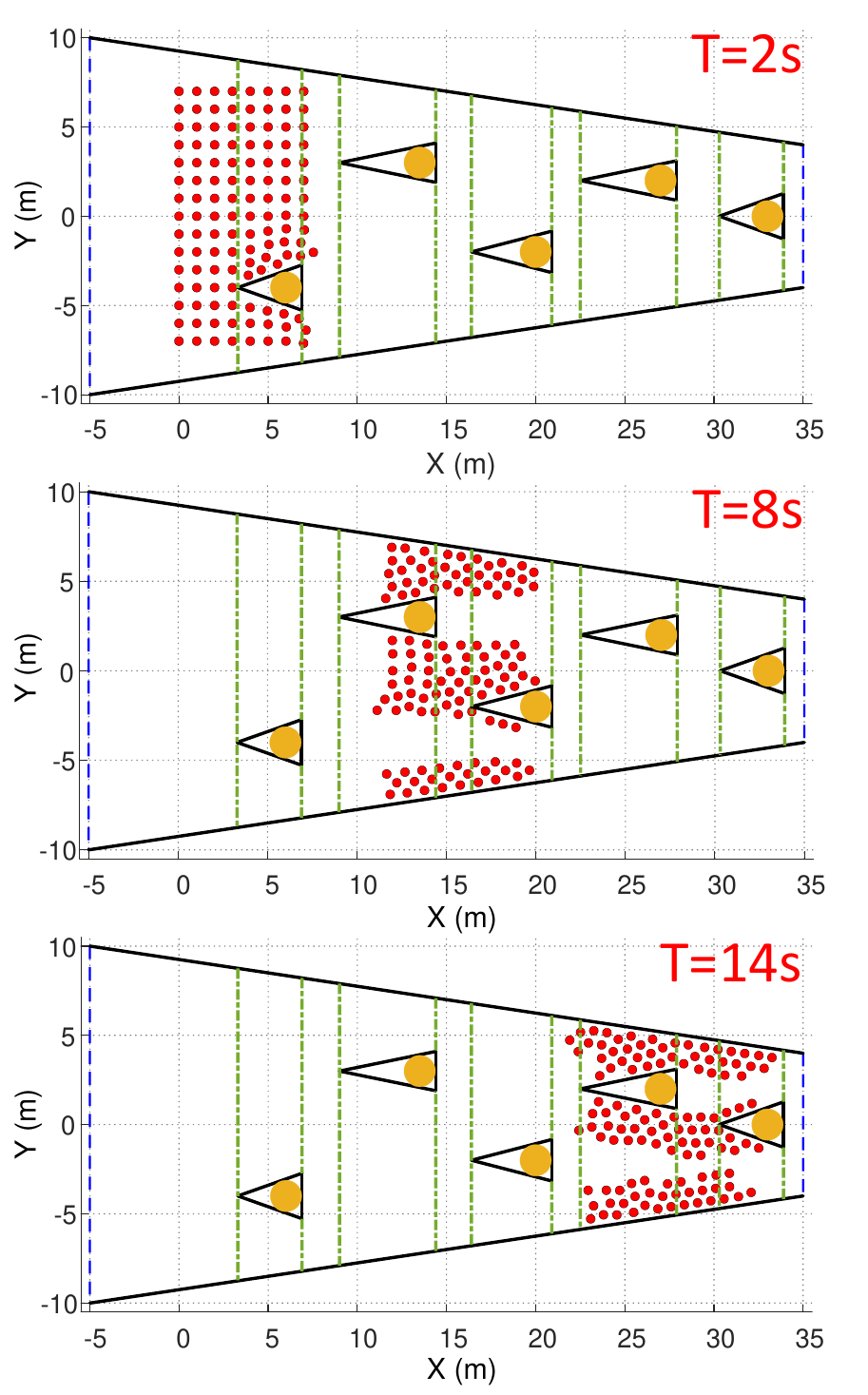}
	\caption{Three snapshots of the numerical simulation. These snapshots show that all robots pass through the trapezoid virtual tube and avoid collision with obstacles successfully.}
	\label{simulation1}
\end{figure}

\begin{figure}[!t]
	\centerline{\includegraphics[width=0.9\columnwidth]{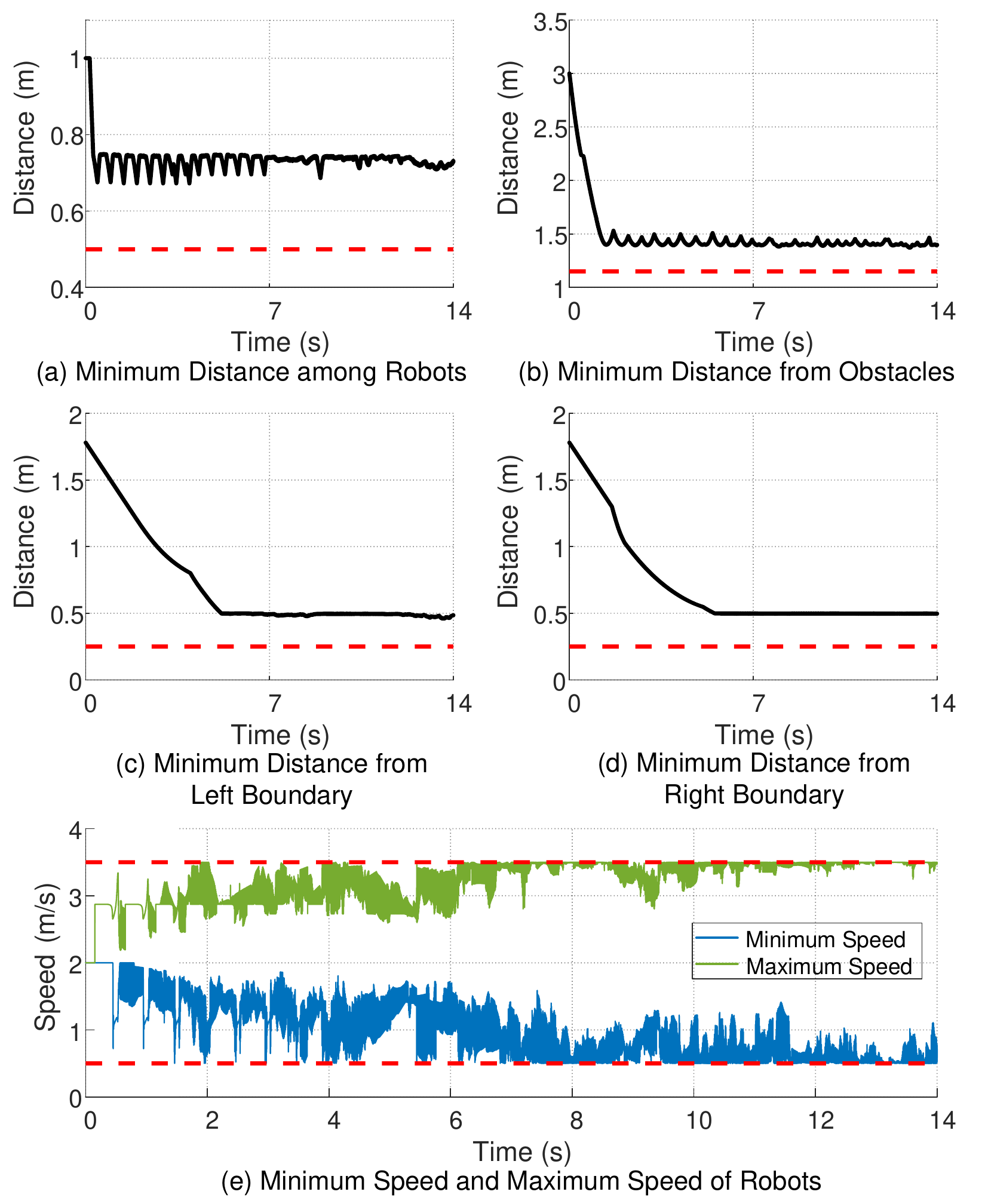}}
	\caption{Some results of the numerical simulation. From plots (a)-(d), it can be seen that all robots keep safe. In plot (e), the reason why two curves chatter is that the robot's model is a single integrator as \eqref{Single}, which allows the sudden change of the robot's velocity.}
	\label{simulation2}
\end{figure}

The validity and feasibility of the proposed method are verified by numerical simulations. The simulation step is set at 0.001s. The scenario considered involves a swarm of $M=120$ robots moving through a predetermined trapezoid virtual tube. All robots satisfy the single integrator model \eqref{Single}. The swarm controller \eqref{ObstacleController} is used to guide the robotic swarm. The trapezoid virtual tube contains five static obstacles. The radii of the obstacles are all set to be $r_{\text{o},k}= 0.9\text{m},k=1,\cdots,5$. At the initial time, all robots with safety radius $r_\text{s} = 0.25\text{m}$ and avoidance radius $r_\text{a}= 0.5\text{m}$ are symmetrically positioned within a rectangular area. The safety area for each robot is represented by a red circle, as shown in Figure \ref{simulation1}. The maximum and the minimum permitted speeds are set to be $v_{\text{min}}=0.5\text{m/s}$, $v_{\text{max}}=3.5\text{m/s}$. To satisfy the constraint \eqref{vminvmax}, the control parameters $v$ in \eqref{u1i} and $v_{\text{max}}^\prime$ in \eqref{ObstacleController} are set to be  $v=2\text{m/s}$, $v_{\text{max}}^\prime=1.5\text{m/s}$. The remaining control parameters are assigned as $k_2=k_3=1$, $\epsilon_{\text{m}}=\epsilon_{\text{t}}=\epsilon_{\text{s}}=10^{-6}$.

The simulation, which lasts 14 seconds, is depicted through three snapshots as shown in Figure \ref{simulation1}. The snapshots show that the robots successfully pass through the trapezoid virtual tube. As depicted in Figure \ref{simulation2}(a), the minimum distance between any two robots is consistently greater than $2r_\text{s}=0.5\text{m}$, indicating that the robotic swarm operates without collisions. In Figure \ref{simulation2}(b), the minimum distance between robots and obstacles is consistently greater than $r_\text{s}+r_{\text{o},k}= 1.15\text{m},k=1,\cdots,5$, indicating that the robots have no collision with obstacles. In addition, Figure \ref{simulation2}(c) and (d) show the minimum distances between the robots and the virtual tube boundary. These minimum distances are always greater than $r_\text{s}=0.25\text{m}$, which implies that all robots keep moving inside the trapezoid virtual tube. During the simulation, the minimum and the maximum speeds of all robots are shown in Figure \ref{simulation2}(e). It is obvious that the constraints in \eqref{vminvmax} are always satisfied. The chattering phenomenon in Figure \ref{simulation2}(e) is caused by the single integrator model, which means that the robot's speed can have a sudden change.

\begin{figure}[!t]
	\centerline{\includegraphics[width=0.9\columnwidth]{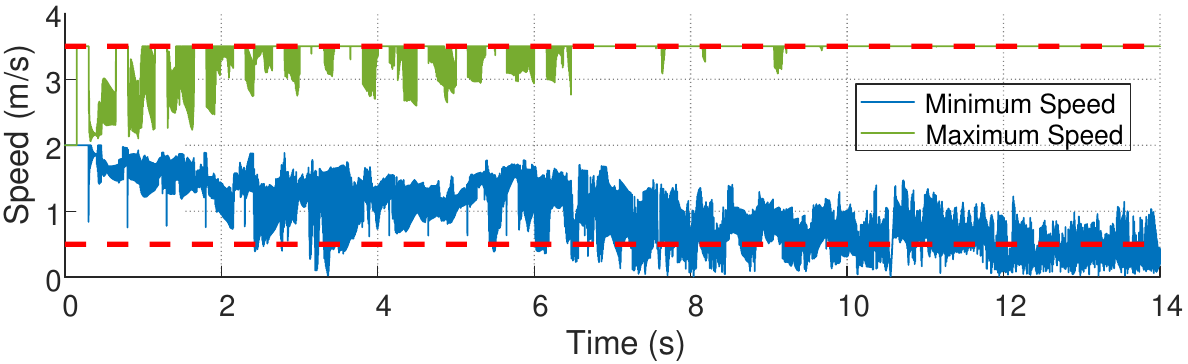}}
	\caption{The minimum speed and maximum speed of robots with the swarm controller proposed in \cite{gao2022distributed}.}
	\label{simulation3}
\end{figure}

\subsection{Comparative Simulation with Our Previous Work}

To show the advantage of the swarm controller proposed in this paper, we make a comparative simulation with the controller designed in our previous work \cite{gao2022distributed}. Using the symbols in \eqref{ModifiedVectorField}, the controller in \cite{gao2022distributed} can be expressed as
\begin{align}
	\mathbf{v}_{\text{c},i}=\text{sat}\left(\mathbf{u}_{1,i}^{\prime\prime}+\mathbf{u}_{2,i}+\mathbf{u}_{3,i}^{\prime}+\mathbf{u}_{4,i}^{\prime},{v_{\text{max}}}\right). \label{Oldcontroller}
\end{align}
Besides, we also use the switching logic proposed in this paper to achieve static obstacle avoidance.

The simulation configurations and parameter settings are the same as those in the last simulation. During the simulation, all robots have a similar motion behavior as shown in Figure \ref{simulation1}. The main difference is the speeds of robots. With the swarm controller \eqref{Oldcontroller}, the minimum and the maximum speeds of robots are shown in Figure \ref{simulation3}. It can be observed that the swarm controller \eqref{Oldcontroller} can only satisfy the maximum speed constraint, and the minimum speed constraint is violated. A direct method is scaling the velocity command to meet the minimum speed constraint. However, when the velocity command equals zero, namely $\mathbf{v}_{\text{c},i}=\mathbf{0}$, such a method is unfeasible as the zero vector has no direction and cannot be scaled. In a word, compared to the controller designed in our previous work \cite{gao2022distributed}, the main advantage of the controller presented in this work is its ability to satisfy the minimum speed constraint.

\subsection{Real Experiment}

\begin{figure}[!t]
	\centering
	\includegraphics[width=0.9\columnwidth]{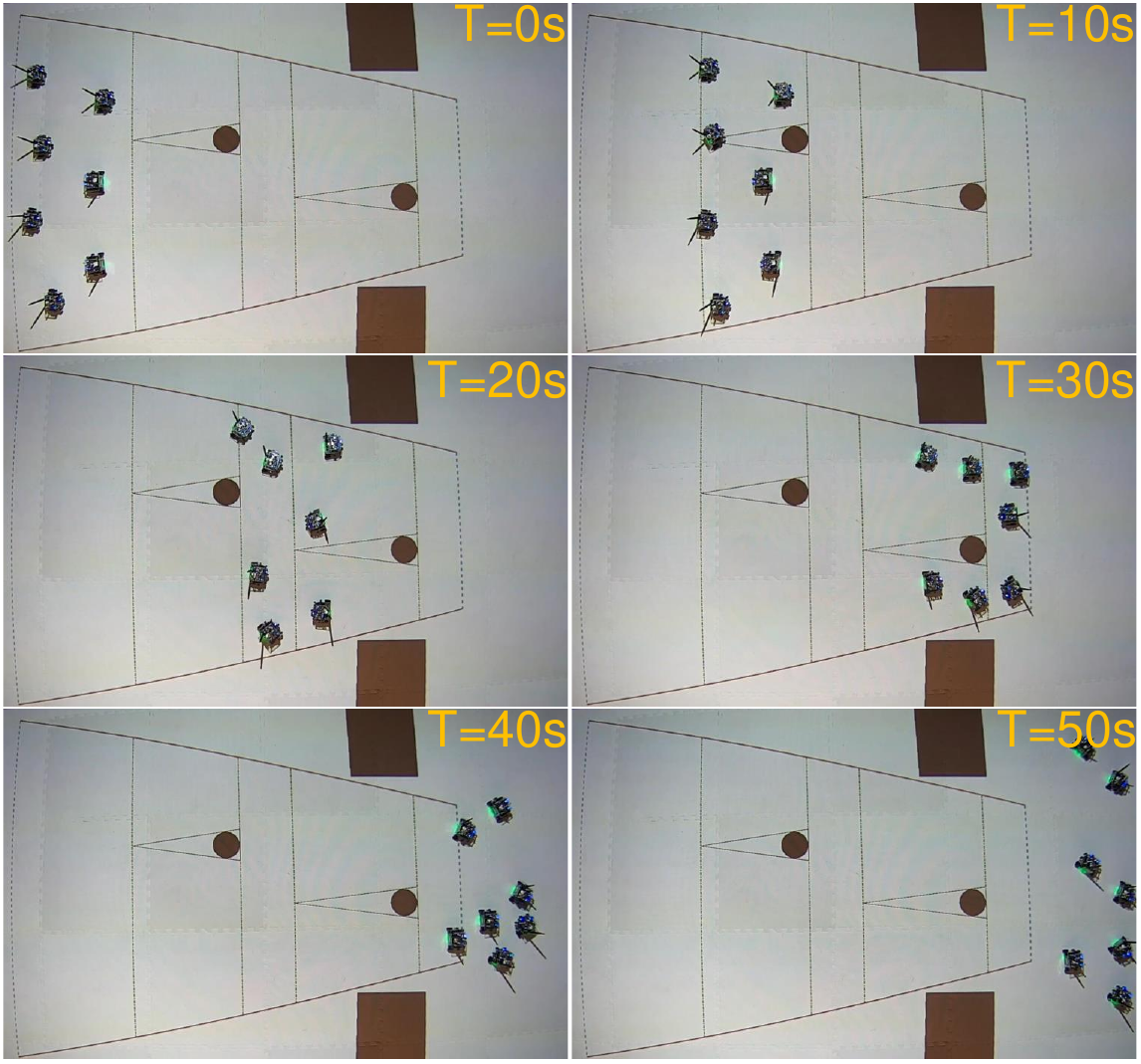}
	\caption{Six snapshots of the real experiment based on Robotarium.}
	\label{ExpSnapshot}
\end{figure}

In this subsection, a real experiment is conducted on the Robotarium, an open-source platform developed by the Georgia Institute of Technology \cite{wilson2020robotarium}.
As shown in Figure \ref{ExpSnapshot}, $M=7$ ground mobile robots move in a predefined trapezoid virtual tube. All ground mobile robots satisfy the nonlinear unicycle model given by
$
	\dot{x}_i=v_{\text{c},i}\cos\theta_i, \dot{y}_i=v_{\text{c},i}\sin\theta_i, \dot{\theta}_i=\omega_{\text{c},i},
$
where $\mathbf{p}_i=\left[x_i\ y_i\right]^\text{T}$ is the position of the $i$th robot, $\theta_i$ is the heading, $v_{\text{c},i}$ and $\omega_{\text{c},i}$ are the linear and angular speed inputs, respectively. To transform the velocity command $\mathbf{v}_{\text{c},i}$ into linear and angular speed inputs $v_{\text{c},i}$, $\omega_{\text{c},i}$, we use a near-identity diffeomorphism (NID) method \cite{olfati2002near}, which defines a control point ahead of the robot with a distance $s>0$. The output feedback control law is given by
\begin{align}
\left[
\begin{matrix}
{v}_{\text{c},i}\\\omega_{\text{c},i}
\end{matrix}
\right]=
\left[
\begin{matrix}
	\cos\theta_i & \sin\theta_i \\
	-\frac{1}{s}\sin\theta_i &  \frac{1}{s}\cos\theta_i
\end{matrix}
\right]\mathbf{v}_{\text{c},i}. \label{NID}
\end{align}
The above control
input transformation can effectively stabilize the robot
with small bounded errors.

In the experiment, there are two static obstacles inside the virtual tube. Each robot reaches its goal point after passing through the trapezoid virtual tube. The radii of the obstacles are both set to be  $r_{\text{o},k}= 0.07\text{m},k=1,2$. All ground robots have $r_\text{s} = 0.075\text{m}$ and $r_\text{a}= 0.125\text{m}$. The controllers \eqref{ObstacleController} and \eqref{NID} are applied for all robots. The maximum and minimum permitted speeds are set to be  $v_{\text{min}}=0.01\text{m/s}$, $v_{\text{max}}=0.09\text{m/s}$. To satisfy \eqref{vminvmax}, the control parameters $v$ in \eqref{u1i} and $v_{\text{max}}^\prime$ in \eqref{ObstacleController} are set to be $v=0.06\text{m/s}$, $v_{\text{max}}^\prime=0.03\text{m/s}$. The remaining control parameters are assigned as $k_2=k_3=1$, $\epsilon_{\text{m}}=\epsilon_{\text{t}}=\epsilon_{\text{s}}=10^{-6},s=0.02\text{m}$.

\begin{figure}[!t]
	\centering
	\includegraphics[width=0.9\columnwidth]{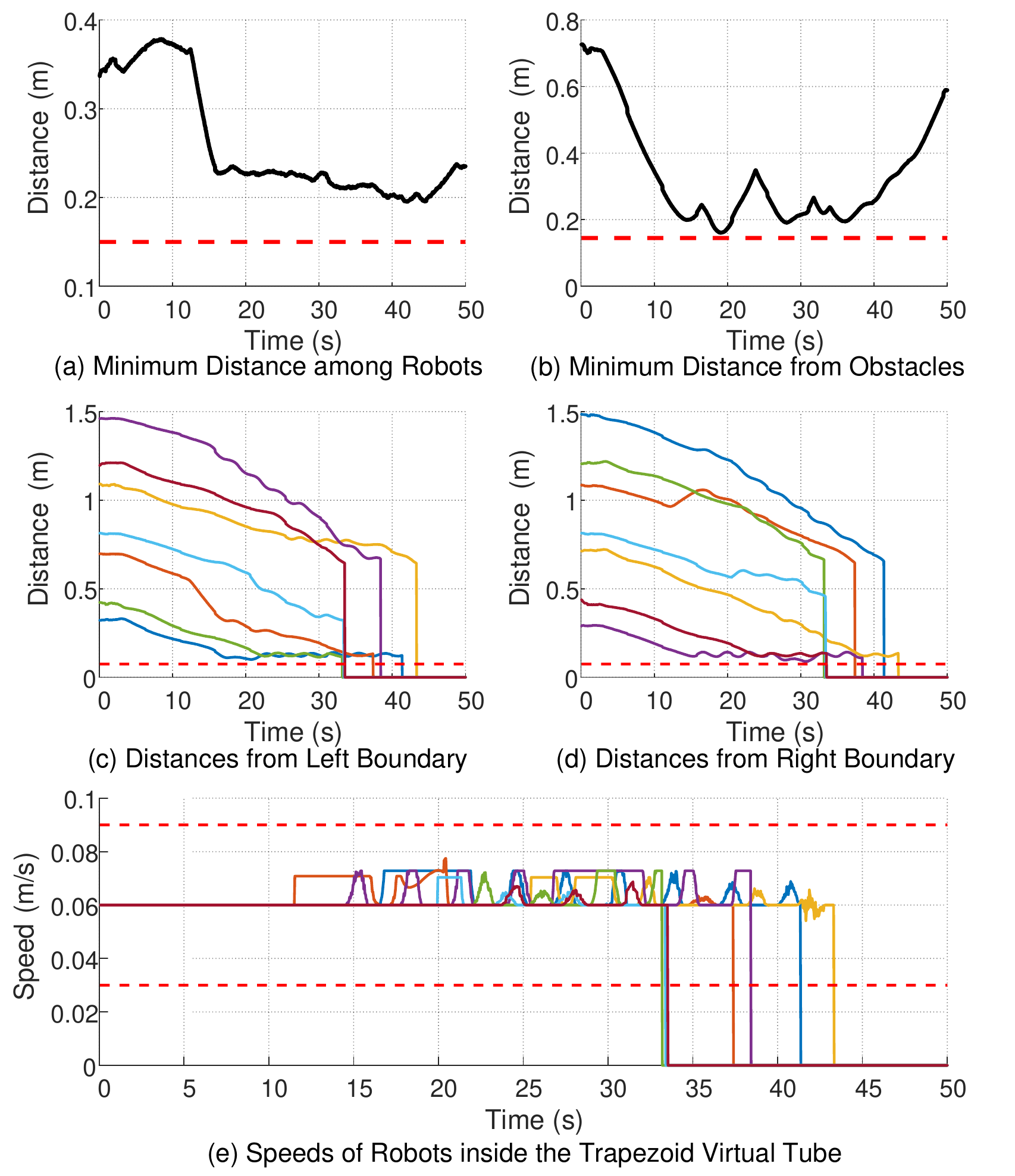}
	\caption{Some results of the real experiment. The curves in plots (c)-(e) approaching zero means that their corresponding robots have passed the finishing line of the trapezoid virtual tube.}
	\label{Expcurve}
\end{figure}

The real experiment lasts 50 seconds and is visualized by six snapshots in Figure \ref{ExpSnapshot}. The experiment shows the successful navigation of all robots inside the trapezoid virtual tube, without any collisions with other robots or obstacles. Furthermore, Figure \ref{Expcurve}(a) demonstrates that the minimum distance between any two robots is always greater than $2r_\text{s}=0.15\text{m}$. Figure \ref{Expcurve}(b) also shows that the minimum distance between robots and obstacles is consistently greater than $r_\text{s}+r_{\text{o},k}= 0.145\text{m},k=1,2$. In addition, Figure \ref{Expcurve}(c) and (d) illustrate the distances between the robots and the virtual tube boundary, which remain constantly greater than $r_\text{s}=0.07\text{m}$ when the robots are inside the trapezoid virtual tube. During the experiment, the speeds of all robots are shown in Figure \ref{Expcurve}(e). It is obvious that the constraints \eqref{vminvmax} are always satisfied when robots are inside the trapezoid virtual tube.

\section{Conclusions}

This paper proposes a solution to the distributed control problem for a robotic swarm to pass through a trapezoid virtual tube with static obstacles and speed constraints. The relationship between the minimum speed constraint and the trapezoid virtual tube is also discussed. The proposed solution is analyzed and proven using carefully designed Lyapunov-like functions. Specifically, \emph{Lemma 1} shows that collision avoidance and keeping within the trapezoid virtual tube can be achieved. \emph{Theorem 1} proves that all robots can pass through the trapezoid virtual tube with no obstacle inside, while \emph{Theorem 2} shows that the proposed method can also handle trapezoid virtual tubes containing static obstacles. The effectiveness and performance of the proposed method are validated by simulations and experiments. The limitation of the proposed method is that the swarm controller only considers speed constraints but not the turning radius constraint, which is important for fixed-wing UAVs. In the future, we are going to investigate this problem. The robustness of our proposed controller will also be the focus.

\appendices

\section{Proof of Lemma 1}
For the robotic swarm inside $\mathcal{T}$, a function is defined as
\begin{equation}
	{V}=\underset{i=1}{\overset{M}{\sum }}\left( V_{\text{f},i}+\frac{1}{2}
	\sum_{j=1,j\neq i}^{M}V_{\text{m},ij}+V_{\text{th},i}+V_{\text{tr},i}\right). \label{V}
\end{equation}
We have $V\geq 0$. According to \eqref{vcipartion}, the relative velocity command of the $i$th robot is shown as   
$
\tilde{\mathbf{v}}_{\text{c},i}=\mathbf{u}_{1,i}^{\prime}+\text{sat}\left(\mathbf{u}_{2,i}+\mathbf{u}_{3,i}+\mathbf{u}_{4,i},{v_{\text{max}}^{\prime}}\right),
$
where $\mathbf{u}_{1,i}^{\prime}=\mathbf{u}_{1,i}-\mathbf{v}^\ast.$ 
According to the single integrator models \eqref{Single} and \eqref{Single1}, the derivative of ${V}$ is shown as 
\begin{align*}
	&{\dot{V}} =\sum_{i=1}^{M}\left(\frac{\partial V_{\text{f},i}}{\partial \tilde{\mathbf{p}}_i}\tilde{\mathbf{v}}_{\text{c},i}-\frac{{1}}{2}\sum_{j\in\mathcal{N}_{\text{m},i}}b_{ij}\tilde{\mathbf{p}}
	_{\text{m,}ij}^{\text{T}}\left( \tilde{\mathbf{v}}_{\text{c},i}+\mathbf{v}^\ast-\tilde{\mathbf{v}}_{\text{c},j}\right.\right. \notag \\
	&\quad\quad\left.\left.-\mathbf{v}^\ast\right) +\frac{\partial V_{\text{th},i}}{\partial \mathbf{p}_i } \left(\tilde{\mathbf{v}}_{\text{c},i}+\mathbf{v}^\ast\right)+\frac{\partial V_{\text{tr},i}}{\partial \mathbf{p}_i } \left(\tilde{\mathbf{v}}_{\text{c},i}+\mathbf{v}^\ast\right)\right) \\
	&\!=\!\sum_{i=1}^{M}\! \left(\!-\!\left(\! \mathbf{u}_{1,i}^{\prime}\!+\!\mathbf{u}_{2,i}\!+\!\mathbf{u}_{3,i}\!+\!\mathbf{u}_{4,i}\right) ^{\text{T}}\!\tilde{\mathbf{v}}_{\text{c},i}\!+\!\left(\!\frac{\partial V_{\text{th},i}}{\partial \mathbf{p}_i }\!+\!\frac{\partial V_{\text{tr},i}}{\partial \mathbf{p}_i }\!\right)\mathbf{v}^\ast\!\right)\!,
\end{align*}
in which $\sum_{j=1,j\neq i}^{M}b_{ij}\tilde{\mathbf{p}}_{\text{m,}ij}=\sum_{j\in\mathcal{N}_{\text{m},i}}b_{ij}\tilde{\mathbf{p}}_{\text{m,}ij}$ is used.
Assume that there are $M_1$ robots within $\mathcal{M}_\mathcal{T}$, $M_2$ robots within $\mathcal{L}_\mathcal{T}$, and $M_3$ robots within $\mathcal{R}_\mathcal{T}$. It is obvious that $\sum_{k=1}^{3}M_k=M$. Then, three functions are defined as 
$
	{V}_k=\underset{i=1}{\overset{M_k}{\sum }}\left( V_{\text{f},i}+\frac{1}{2}
	\sum_{j=1,j\neq i}^{M}V_{\text{m},ij}+V_{\text{th},i}+V_{\text{tr},i}\right) ,
$
where $k=1,2,3$. We have $V=\sum_{k=1}^{3}{V}_k$ and $\dot{V}=\sum_{k=1}^{3}\dot{V}_k$. Here the reference velocity command $\mathbf{v}^\ast$ is selected as $\mathbf{v}^\ast=v\mathbf{n}_\text{c}$. According to \eqref{DirL2}, it is obtained that 
$
\left(\partial V_{\text{th},i}/\partial \mathbf{p}_i+\partial V_{\text{tr},i}/\partial \mathbf{p}_i\right)\mathbf{v}^\ast\leq0. 
$
Then we have
$
{\dot{V}} \leq\sum_{i=1}^{M}-\left( \mathbf{u}_{1,i}^{\prime}+\mathbf{u}_{2,i}+\mathbf{u}_{3,i}+\mathbf{u}_{4,i}\right) ^{\text{T}}\tilde{\mathbf{v}}_{\text{c},i}
$
and
$
{\dot{V}}_k \leq\sum_{i=1}^{M_k}-\left( \mathbf{u}_{1,i}^{\prime}+\mathbf{u}_{2,i}+\mathbf{u}_{3,i}+\mathbf{u}_{4,i}\right) ^{\text{T}}\tilde{\mathbf{v}}_{\text{c},i},
$
where $k=1,2,3$.
For any robot within $\mathcal{M}_\mathcal{T}$, we have $\mathbf{u}_{1,i}=\mathbf{v}^\ast=v\mathbf{n}_\text{c}$ and $\mathbf{u}_{1,i}^{\prime}=\mathbf{0}$, where $i=1,\cdots,M_1$.  Let $\mathbf{u}_{\text{r},i}=\mathbf{u}_{2,i}+\mathbf{u}_{3,i}+\mathbf{u}_{4,i}$. Then we have 
$
	{\dot{V}}_1 \leq\sum_{i=1}^{M_1} -\mathbf{u}_{\text{r},i} ^{\text{T}}\text{sat}\left(\mathbf{u}_{\text{r},i},{v_{\text{max}}^{\prime}}\right)\leq \sum_{i=1}^{M_1}-\kappa_{v_{\text{max}}^{\prime}}\mathbf{u}_{\text{r},i}^{\text{T}}\mathbf{u}_{\text{r},i}\leq 0.
$
For any robot within $\mathcal{L}_\mathcal{T}$, there exist $\mathbf{u}_{1,i}=v\mathbf{t}_\text{h}$ and $\mathbf{u}_{1,i}^{\prime}=v\left(\mathbf{t}_\text{h}-\mathbf{n}_\text{c}\right)$, where $i=1,\cdots,M_2$. Then we have 
\begin{align*}
	&{\dot{V}}_2 \leq\sum_{i=1}^{M_2}-\left(\mathbf{u}_{1,i}^{\prime}+ \mathbf{u}_{\text{r},i}\right) ^{\text{T}}\left(\mathbf{u}_{1,i}^{\prime}+\text{sat}\left(\mathbf{u}_{\text{r},i},{v_{\text{max}}^{\prime}}\right)\right)\\
	&\leq \sum_{i=1}^{M_2}\!-\!\left(\left\Vert\mathbf{u}_{1,i}^{\prime}\right\Vert^2\!-\!\left(1\!+\!\kappa_{v_{\text{max}}^{\prime}}\right)\left\Vert\mathbf{u}_{1,i}^{\prime}\right\Vert\left\Vert\mathbf{u}_{\text{r},i}\right\Vert\!+\!\kappa_{v_{\text{max}}^{\prime}}\left\Vert\mathbf{u}_{\text{r},i}\right\Vert^2\right).
\end{align*}
When $\left\Vert\mathbf{u}_{\text{r},i}\right\Vert\leq{v_{\text{max}}^{\prime}}$, it has $\kappa_{v_{\text{max}}^{\prime}}=1$, which results in 
$
{\dot{V}}_2 \leq \sum_{i=1}^{M_2}-\left(\left\Vert\mathbf{u}_{1,i}^{\prime}\right\Vert-\left\Vert\mathbf{u}_{\text{r},i}\right\Vert\right)^2\leq 0.
$
When $\left\Vert\mathbf{u}_{\text{r},i}\right\Vert>{v_{\text{max}}^{\prime}}$, it has $\kappa_{v_{\text{max}}^{\prime}}=\frac{v_{\text{max}}^{\prime}}{\left\Vert\mathbf{u}_{\text{r},i}\right\Vert}$. Besides, according to the constraint \eqref{Constraint2}, we have $\left\Vert\mathbf{u}_{1,i}^{\prime}\right\Vert<v_{\text{max}}^{\prime}$, which results in $\left\Vert\mathbf{u}_{\text{r},i}\right\Vert>{v_{\text{max}}^{\prime}}>\left\Vert\mathbf{u}_{1,i}^{\prime}\right\Vert$. Then we have
\begin{align*}
	{\dot{V}}_2 &\leq \sum_{i=1}^{M_2}\!-\!\left(\left\Vert\mathbf{u}_{1,i}^{\prime}\right\Vert^2\!-\!\left(1\!+\!\kappa_{v_{\text{max}}^{\prime}}\right)\left\Vert\mathbf{u}_{1,i}^{\prime}\right\Vert\left\Vert\mathbf{u}_{\text{r},i}\right\Vert\!+\!\kappa_{v_{\text{max}}^{\prime}}\left\Vert\mathbf{u}_{\text{r},i}\right\Vert^2\right)\\
	&\leq \sum_{i=1}^{M_2}\!-\!\left(\left\Vert\mathbf{u}_{1,i}^{\prime}\right\Vert-{v_{\text{max}}^{\prime}}\right)\left(\left\Vert\mathbf{u}_{1,i}^{\prime}\right\Vert-\left\Vert\mathbf{u}_{\text{r},i}\right\Vert\right) \leq 0.
\end{align*}
In a word, ${\dot{V}}_2 \leq 0$ is always satisfied. Similarly, we can get ${\dot{V}}_3 \leq 0$ according to the constraint \eqref{Constraint2}. By using the velocity command \eqref{VectorField} for all robots, $\dot{V}$ satisfies $\dot{V}=\dot{V}_1+\dot{V}_2+\dot{V}_3\leq 0$. 

Then, the reasons why these robots can avoid conflict with each other and stay within the trapezoid virtual tube can be found in the \emph{Lemma 2} of \cite{gao2022distributed}, so we will not go into much detail here. $\square $

\section{Proof of Theorem 1}
Based on \textit{Lemma 1}, it is guaranteed that all robots can avoid conflict with each other and remain inside the trapezoid virtual tube. Next, we will explain why all robots can approach the finishing line $\left[ {{\mathbf{p}}_{\text{r,}1},{\mathbf{p}}_{\text{h,}1}}\right]$. We assume that the $1$st robot is positioned at the front of the swarm. In other words, it is the closest to the finishing line. We denote the distance between the $1$st robot and the finishing line as
$
d{_{\text{f,}1}}=\mathbf{n}_{\text{c}}^{\text{T}}\left({\mathbf{p}}_{\text{r,}1}-\mathbf{p}_{1}\right).
$
When $\mathbf{p}_{1} \in \mathcal{T}$, we have $d{_{\text{f,}1}}\geq 0$. Then, a Lyapunov function is defined as 
$
V_{\text{fl},1}=\frac{1}{2}d_{\text{f,}1}^2. 
$
It is obvious that $V_{\text{fl},1}\geq0$. When $d{_{\text{f,}1}}=0$, namely the 1st robot is on the finishing line, it has $V_{\text{fl},1}=0$. Then, the derivative of $V_{\text{fl},1}$ is shown as 
$
	\dot{V}_{\text{fl},1}=-d_{\text{f,}1}\mathbf{n}_{\text{c}}^{\text{T}}\left(\mathbf{u}_{1,1}+\kappa_{v_{\text{max}}^{\prime}}\left( \mathbf{u}_{2,1}+\mathbf{u}_{3,1}+\mathbf{u}_{4,1}\right)\right).
$
Given that the $1$st robot is ahead, it can be stated that $\mathbf{n}_{\text{c}}^{\text{T}}\mathbf{\tilde{p}}_{\text{m,}1j}\geq 0$, with the equality being true only when the $j$th robot is as ahead as the $1$st robot.
Then we have
$
-d_{\text{f,}1}\kappa_{v_{\text{max}}^{\prime}}\mathbf{n}_{\text{c}}^{\text{T}}\mathbf{u}_{2,1} \leq 0.
$
Besides, according to the relationship \eqref{DirL2}, we have
$
-d_{\text{f,}1}\kappa_{v_{\text{max}}^{\prime}}\mathbf{n}_{\text{c}}^{\text{T}}\left(\mathbf{u}_{3,1}+\mathbf{u}_{4,1}\right)\leq0.
$
Then we have 
$
\dot{V}_{\text{fl},1}\leq -d_{\text{f,}1}\mathbf{n}_{\text{c}}^{\text{T}}\mathbf{u}_{1,1}.
$
As there exist $\mathbf{n}_{\text{c}}^{\text{T}}\mathbf{n}_{\text{c}} >0$, $\mathbf{t}_{\text{h}}^{\text{T}}\mathbf{n}_{\text{c}} >0$,  $\mathbf{t}_{\text{r}}^{\text{T}}\mathbf{n}_{\text{c}} >0$, we have
$
\dot{V}_{\text{fl},1}< 0
$
according to \eqref{u1i}.
Given the initial condition $d{_{\text{f,}1}}
\left( 0\right) >{0}$ from \textit{Assumption 1}, and the fact that the system is continuous, we can guarantee the existence of a time $t_{11}>0$, such that $d{_{\text{f,}1}}\left(t\right)\leq {\epsilon }_{\text{0}}$ at $t\geq t_{11}$, where $\epsilon_0>0$. At $t_{11}$, the $1$st robot is removed from the trapezoid virtual tube in accordance with \textit{Assumption 2}. The remaining problem is to analyze the $M-1$ robots. This analysis can be repeated to complete the proof.
$\square $ 

\section{Problem of Introducing a Potential-based Obstacle Avoidance Term}
For the $i$th robot avoiding collision with the $k$th static obstacle, a barrier function can be  defined as 
$
	V_{\text{o},ik}=V_{\text{n}}(k_4,\left \Vert \mathbf{p}_i-\mathbf{p}_{\text{o},k}\right \Vert,r_{\text{o},k}+r_{\text{s}},r_{\text{o},k}+r_{\text{a}},\epsilon_{\text{o}},\epsilon_{\text{s}}),
$
where $k_4,\epsilon _{\text{o}},\epsilon _{\text{s}}>0$. By introducing a potential-based obstacle avoidance term, the swarm controller \eqref{VectorField} becomes
$
	\mathbf{v}_{\text{c},i}=\mathbf{u}_{1,i}+\text{sat}\left(\mathbf{u}_{2,i}+\mathbf{u}_{3,i}+\mathbf{u}_{4,i}+\mathbf{u}_{5,i},{v_{\text{max}}^{\prime}}\right),
$
in which the obstacle avoidance term $\mathbf{u}_{5,i}$ is written as 
$
	\mathbf{u}_{5,i}=\sum_{k=1}^{P}-\left(\frac{\partial V_{\text{o},ik}}{\partial \mathbf{p}_i }\right)^{\text{T}}=\sum_{k=1}^{P}
	-\frac{\partial V_{\text{o},ik}}{\partial \left \Vert \mathbf{p}_i-\mathbf{p}_{\text{o},k}\right \Vert }\frac{\mathbf{p}_i-\mathbf{p}_{\text{o},k}}{\left \Vert \mathbf{p}_i-\mathbf{p}_{\text{o},k}\right \Vert}.
$
Based on \eqref{V}, a similar function is designed as ${V}^\prime=V+\sum_{i=1}^{M}\sum_{k=1}^{P}V_{\text{o},ik}$. We have $V^\prime\geq 0$. Then $\dot{V}^\prime$ is shown as
\begin{align*}
	{\dot{V}^\prime} =\sum_{i=1}^{M} \left(-\left( \mathbf{u}_{1,i}^{\prime}+\mathbf{u}_{2,i}+\mathbf{u}_{3,i}+\mathbf{u}_{4,i}+\mathbf{u}_{5,i}\right) ^{\text{T}}\tilde{\mathbf{v}}_{\text{c},i}\right. \\
	\left.+\left(\frac{\partial V_{\text{th},i}}{\partial \mathbf{p}_i }+\frac{\partial V_{\text{tr},i}}{\partial \mathbf{p}_i }+\sum_{k=1}^{P}\frac{\partial V_{\text{o},ik}}{\partial \mathbf{p}_i }\right)\mathbf{v}^\ast\right).
\end{align*}
If $\frac{\partial V_{\text{o},ik}}{\partial \mathbf{p}_i}\mathbf{v}^\ast\leq0,i=1,\cdots,M,k=1,\cdots,P$ is satisfied, it has $\dot{V}^\prime\leq0$ according to the analysis in the proof of \emph{Lemma 1}. Then we have
$\frac{\partial V_{\text{o},ik}}{\partial \mathbf{p}_i}\mathbf{v}^\ast=\frac{\partial V_{\text{o},ik}}{\partial \left \Vert \mathbf{p}_i-\mathbf{p}_{\text{o},k}\right \Vert }\frac{\left(\mathbf{p}_i-\mathbf{p}_{\text{o},k}\right)^{\text{T}}}{\left \Vert \mathbf{p}_i-\mathbf{p}_{\text{o},k}\right \Vert}\mathbf{v}^\ast$, 
where $\frac{\partial V_{\text{o},ik}}{\partial \left \Vert \mathbf{p}_i-\mathbf{p}_{\text{o},k}\right \Vert }\leq0$ according to the first property of the nominal Lyapunov-like barrier function $V_{\text{n}}$.
However, if $\mathbf{v}^\ast$ is chosen as $\mathbf{v}^\ast=v\mathbf{n}_\text{c}$, we have $\left(\mathbf{p}_i-\mathbf{p}_{\text{o},k}\right)^{\text{T}}\mathbf{v}^\ast<0$ when the $i$th robot is behind the $k$th obstacle. Such a phenomenon is similar to the dynamic obstacle avoidance problem of the APF method. With this phenomenon, we have $\frac{\partial V_{\text{o},ik}}{\partial \mathbf{p}_i}\mathbf{v}^\ast\geq0$, and $\dot{V}^\prime\leq 0$ may not be satisfied. Hence, the safety and convergence proof cannot be obtained.


\bibliographystyle{IEEEtran}
\bibliography{FW}

\begin{thebibliography}{10}
\providecommand{\url}[1]{#1}
\csname url@samestyle\endcsname
\providecommand{\newblock}{\relax}
\providecommand{\bibinfo}[2]{#2}
\providecommand{\BIBentrySTDinterwordspacing}{\spaceskip=0pt\relax}
\providecommand{\BIBentryALTinterwordstretchfactor}{4}
\providecommand{\BIBentryALTinterwordspacing}{\spaceskip=\fontdimen2\font plus
\BIBentryALTinterwordstretchfactor\fontdimen3\font minus
  \fontdimen4\font\relax}
\providecommand{\BIBforeignlanguage}[2]{{%
\expandafter\ifx\csname l@#1\endcsname\relax
\typeout{** WARNING: IEEEtran.bst: No hyphenation pattern has been}%
\typeout{** loaded for the language `#1'. Using the pattern for}%
\typeout{** the default language instead.}%
\else
\language=\csname l@#1\endcsname
\fi
#2}}
\providecommand{\BIBdecl}{\relax}
\BIBdecl

\bibitem{zhou2022swarm}
X.~Zhou, X.~Wen, Z.~Wang, Y.~Gao, H.~Li, Q.~Wang, T.~Yang, H.~Lu, Y.~Cao,
  C.~Xu, and F.~Gao, ``Swarm of micro flying robots in the wild,''
  \emph{Science Robotics}, vol.~7, no.~66, p. eabm5954, 2022.

\bibitem{soria2021predictive}
E.~Soria, F.~Schiano, and D.~Floreano, ``Predictive control of aerial swarms in
  cluttered environments,'' \emph{Nature Machine Intelligence}, vol.~3, no.~6,
  pp. 545--554, 2021.

\bibitem{hu2022decentralized}
Y.~Hu, J.~Fu, and G.~Wen, ``Decentralized robust collision avoidance for
  cooperative multirobot systems: A gaussian process-based control barrier
  function approach,'' \emph{IEEE Transactions on Control of Network Systems},
  vol.~10, no.~2, pp. 706--717, 2023.

\bibitem{lavaei2022formal}
A.~Lavaei, L.~Di~Lillo, A.~Censi, and E.~Frazzoli, ``Formal estimation of
  collision risks for autonomous vehicles: A compositional data-driven
  approach,'' \emph{IEEE Transactions on Control of Network Systems}, vol.~10,
  no.~1, pp. 407--418, 2022.

\bibitem{zhao2019bearing}
S.~Zhao and D.~Zelazo, ``Bearing rigidity theory and its applications for
  control and estimation of network systems: Life beyond distance rigidity,''
  \emph{IEEE Control Systems Magazine}, vol.~39, no.~2, pp. 66--83, 2019.

\bibitem{zhang2021robust}
Q.~Zhang and H.~H. Liu, ``Robust nonlinear close formation control of multiple
  fixed-wing aircraft,'' \emph{Journal of Guidance, Control, and Dynamics},
  vol.~44, no.~3, pp. 572--586, 2021.

\bibitem{xu2020affine}
Y.~Xu, S.~Zhao, D.~Luo, and Y.~You, ``Affine formation maneuver control of
  high-order multi-agent systems over directed networks,'' \emph{Automatica},
  vol. 118, p. 109004, 2020.

\bibitem{zhou2021ego}
X.~Zhou, J.~Zhu, H.~Zhou, C.~Xu, and F.~Gao, ``Ego-swarm: A fully autonomous
  and decentralized quadrotor swarm system in cluttered environments,'' in
  \emph{2021 IEEE International Conference on Robotics and Automation
  (ICRA)}.\hskip 1em plus 0.5em minus 0.4em\relax IEEE, 2021, pp. 4101--4107.

\bibitem{park2020online}
J.~Park and H.~J. Kim, ``Online trajectory planning for multiple quadrotors in
  dynamic environments using relative safe flight corridor,'' \emph{IEEE
  Robotics and Automation Letters}, vol.~6, no.~2, pp. 659--666, 2020.

\bibitem{park2022online}
J.~Park, D.~Kim, G.~C. Kim, D.~Oh, and H.~J. Kim, ``Online distributed
  trajectory planning for quadrotor swarm with feasibility guarantee using
  linear safe corridor,'' \emph{IEEE Robotics and Automation Letters}, vol.~7,
  no.~2, pp. 4869--4876, 2022.

\bibitem{toumieh2022decentralized}
C.~Toumieh and A.~Lambert, ``Decentralized multi-agent planning using model
  predictive control and time-aware safe corridors,'' \emph{IEEE Robotics and
  Automation Letters}, vol.~7, no.~4, pp. 11\,110--11\,117, 2022.

\bibitem{kushleyev2013towards}
A.~Kushleyev, D.~Mellinger, C.~Powers, and V.~Kumar, ``Towards a swarm of agile
  micro quadrotors,'' \emph{Autonomous Robots}, vol.~35, no.~4, pp. 287--300,
  2013.

\bibitem{panagou2015distributed}
D.~Panagou, D.~M. Stipanovi{\'c}, and P.~G. Voulgaris, ``Distributed
  coordination control for multi-robot networks using {Lyapunov}-like barrier
  functions,'' \emph{IEEE Transactions on Automatic Control}, vol.~61, no.~3,
  pp. 617--632, 2015.

\bibitem{wang2017safety}
L.~Wang, A.~D. Ames, and M.~Egerstedt, ``Safety barrier certificates for
  collisions-free multirobot systems,'' \emph{IEEE Transactions on Robotics},
  vol.~33, no.~3, pp. 661--674, 2017.

\bibitem{xie2022distributed}
S.~Xie, J.~Hu, P.~Bhowmick, Z.~Ding, and F.~Arvin, ``Distributed motion
  planning for safe autonomous vehicle overtaking via artificial potential
  field,'' \emph{IEEE Transactions on Intelligent Transportation Systems},
  vol.~23, no.~11, pp. 21\,531--21\,547, 2022.

\bibitem{chiang2015path}
H.-T. Chiang, N.~Malone, K.~Lesser, M.~Oishi, and L.~Tapia, ``Path-guided
  artificial potential fields with stochastic reachable sets for motion
  planning in highly dynamic environments,'' in \emph{2015 IEEE International
  Conference on Robotics and Automation (ICRA)}.\hskip 1em plus 0.5em minus
  0.4em\relax IEEE, 2015, pp. 2347--2354.

\bibitem{sharma2022priority}
A.~Sharma and N.~K. Sinha, ``Priority tagged artificial potential field swarm
  formation control through narrow corridors,'' \emph{IFAC-PapersOnLine},
  vol.~55, no.~22, pp. 358--362, 2022.

\bibitem{liu2018brain}
Y.~Liu, Z.~Li, T.~Zhang, and S.~Zhao, ``Brain--robot interface-based navigation
  control of a mobile robot in corridor environments,'' \emph{IEEE Transactions
  on Systems, Man, and Cybernetics: Systems}, vol.~50, no.~8, pp. 3047--3058,
  2018.

\bibitem{guerra2016avoiding}
M.~Guerra, D.~Efimov, G.~Zheng, and W.~Perruquetti, ``Avoiding local minima in
  the potential field method using input-to-state stability,'' \emph{Control
  Engineering Practice}, vol.~55, pp. 174--184, 2016.

\bibitem{kim1992real}
J.~O. Kim and P.~Khosla, ``Real-time obstacle avoidance using harmonic
  potential functions,'' Ph.D. dissertation, Carnegie Mellon University, 1992.

\bibitem{afzal2022model}
W.~Afzal and A.~A. Masoud, ``Model-based navigation control for
  communication-aware motion using harmonic potential fields,'' \emph{IEEE
  Transactions on Automation Science and Engineering}, vol.~20, no.~4, pp.
  2636--2654, 2022.

\bibitem{rimon1990exact}
E.~Rimon, ``Exact robot navigation using artificial potential functions,''
  Ph.D. dissertation, Yale University, 1990.

\bibitem{panagou2014motion}
D.~Panagou, ``Motion planning and collision avoidance using navigation vector
  fields,'' in \emph{2014 IEEE International Conference on Robotics and
  Automation (ICRA)}.\hskip 1em plus 0.5em minus 0.4em\relax IEEE, 2014, pp.
  2513--2518.

\bibitem{breeden2023robust}
J.~Breeden and D.~Panagou, ``Robust control barrier functions under high
  relative degree and input constraints for satellite trajectories,''
  \emph{Automatica}, vol. 155, p. 111109, 2023.

\bibitem{ibuki2020optimization}
T.~Ibuki, S.~Wilson, J.~Yamauchi, M.~Fujita, and M.~Egerstedt,
  ``Optimization-based distributed flocking control for multiple rigid
  bodies,'' \emph{IEEE Robotics and Automation Letters}, vol.~5, no.~2, pp.
  1891--1898, 2020.

\bibitem{gao2022distributed}
Y.~Gao, C.~Bai, and Q.~Quan, ``Distributed control for a multiagent system to
  pass through a connected quadrangle virtual tube,'' \emph{IEEE Transactions
  on Control of Network Systems}, vol.~10, no.~2, pp. 693--705, 2023.

\bibitem{wang2020neural}
J.~Wang, W.~Chi, C.~Li, C.~Wang, and M.~Q.-H. Meng, ``Neural {RRT*}:
  Learning-based optimal path planning,'' \emph{IEEE Transactions on Automation
  Science and Engineering}, vol.~17, no.~4, pp. 1748--1758, 2020.

\bibitem{mattamala2022efficient}
M.~Mattamala, N.~Chebrolu, and M.~Fallon, ``An efficient locally reactive
  controller for safe navigation in visual teach and repeat missions,''
  \emph{IEEE Robotics and Automation Letters}, vol.~7, no.~2, pp. 2353--2360,
  2022.

\bibitem{wang2019motion}
Y.~Wang, M.~Shan, and D.~Wang, ``Motion capability analysis for multiple
  fixed-wing {UAV} formations with speed and heading rate constraints,''
  \emph{IEEE Transactions on Control of Network Systems}, vol.~7, no.~2, pp.
  977--989, 2019.

\bibitem{gao2023non}
Y.~Gao, C.~Bai, R.~Fu, and Q.~Quan, ``A non-potential orthogonal vector field
  method for more efficient robot navigation and control,'' \emph{Robotics and
  Autonomous Systems}, vol. 159, p. 104291, 2023.

\bibitem{schmitt2014collision}
L.~Schmitt and W.~Fichter, ``Collision-avoidance framework for small fixed-wing
  unmanned aerial vehicles,'' \emph{Journal of Guidance, Control, and
  Dynamics}, vol.~37, no.~4, pp. 1323--1329, 2014.

\bibitem{xue2020tiny}
F.~Xue, A.~Ming, and Y.~Zhou, ``Tiny obstacle discovery by occlusion-aware
  multilayer regression,'' \emph{IEEE Transactions on Image Processing},
  vol.~29, pp. 9373--9386, 2020.

\bibitem{wilson2020robotarium}
S.~Wilson, P.~Glotfelter, L.~Wang, S.~Mayya, G.~Notomista, M.~Mote, and
  M.~Egerstedt, ``The robotarium: Globally impactful opportunities, challenges,
  and lessons learned in remote-access, distributed control of multirobot
  systems,'' \emph{IEEE Control Systems Magazine}, vol.~40, no.~1, pp. 26--44,
  2020.

\bibitem{olfati2002near}
R.~Olfati-Saber, ``Near-identity diffeomorphisms and exponential
  $\epsilon$-tracking and $\epsilon$-stabilization of first-order nonholonomic
  {$SE(2)$} vehicles,'' in \emph{2002 American Control Conference (ACC)}.\hskip
  1em plus 0.5em minus 0.4em\relax IEEE, 2002, pp. 4690--4695.

\end{thebibliography}

\begin{IEEEbiography}[{\includegraphics[width=1in,height=1.25in,clip,keepaspectratio]{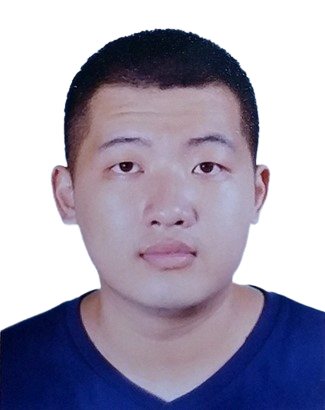}}]
{Yan Gao} received the B.S. degree in detection guidance and control techniques from Harbin Institute of Technology, Harbin, China in 2017, and the Ph.D. degree in control science and engineering from Beihang University, Beijing, China in 2023. He is currently a Lecturer with the School of Control Science and Engineering, Tiangong University, Tianjin, China. His research interests include multi-agent systems, swarm intelligence, virtual tube control and multicopter control.
\end{IEEEbiography}

\begin{IEEEbiography}[{\includegraphics[width=1in,height=1.25in,clip,keepaspectratio]{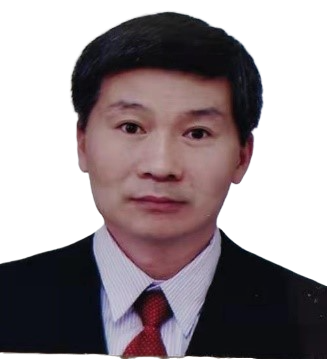}}]{Chenggang Bai} received the MSc degree in statistics from the Nanjing University of Aeronautics and Astronautics, China, in 1990, and the PhD degree in control theory and control engineering from Zhejiang University, China. In November 2001, he joined the faculty of the School of Automation Science and Electrical Engineering, Beihang University, China, where he has been a professor since July 2009. His research interests include reliable flight control, software reliability, and software testing.
\end{IEEEbiography}

\begin{IEEEbiography}[{\includegraphics[width=1in,height=1.25in,clip,keepaspectratio]{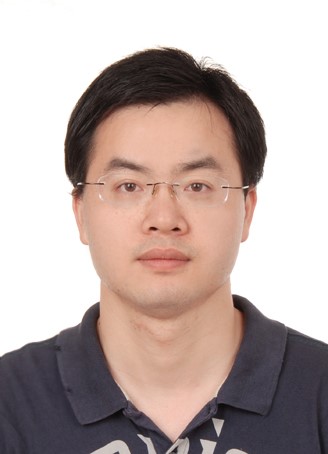}}] {Quan Quan} (M’19) received the B.S. and Ph.D. degrees in
	control science and engineering from Beihang University, Beijing, China, in 2004, and 2010, respectively.
	Since 2022, he has been a Professor with Beihang
	University in control science and engineering, where he
	is currently with the School of Automation Science and
	Electrical Engineering. His research interests include
	reliable flight control, swarm intelligence, vision-based
	navigation, and health assessment.
\end{IEEEbiography}

\end{document}